\documentclass[10pt]{article} 
\usepackage[accepted]{rlc}

\usepackage{amssymb}            
\usepackage{mathtools}          
\usepackage{mathrsfs}           
\usepackage{graphicx}           
\usepackage{subcaption}         
\usepackage[space]{grffile}     
\usepackage{url}                

\usepackage{algorithm}
\usepackage{algcompatible}
\usepackage{amssymb}
\usepackage{amsmath}
\graphicspath{{figures/}}

\DeclareMathOperator*{\argmax}{arg\,max}

\newtheorem{remark}{Remark}
\newtheorem{theorem}{Theorem}
\newtheorem{definition}{Definition}
\newtheorem{lemma}{Lemma}

\title{Improving Thompson Sampling via Information Relaxation for Budgeted Multi-armed Bandits}

\author{Woojin Jeong  \\
    wjddnwls5824@kaist.ac.kr \\
    Department of Industrial \& Systems Engineering\\
    KAIST, Daejeon, Republic of Korea
    \And
    Seungki Min \\
    skmin@kaist.ac.kr\\
    Department of Industrial \& Systems Engineering\\
    KAIST, Daejeon, Republic of Korea}

\begin{document}

\maketitle

\begin{abstract}
We consider a Bayesian budgeted multi-armed bandit problem, in which 
each arm consumes a different amount of resources when selected and there is a budget constraint on the total amount of resources that can be used.
Budgeted Thompson Sampling (BTS) offers a very effective heuristic to this problem, but its arm-selection rule does not take into account the remaining budget information.
We adopt \textit{Information Relaxation Sampling} framework that generalizes Thompson Sampling for classical $K$-armed bandit problems, and propose a series of algorithms that are randomized like BTS but more carefully optimize their decisions with respect to the budget constraint.
In a one-to-one correspondence with these algorithms, a series of performance benchmarks that improve the conventional benchmark are also suggested.
Our theoretical analysis and simulation results show that our algorithms (and our benchmarks) make incremental improvements over BTS (respectively, the conventional benchmark) across various settings including a real-world example.
\end{abstract}

\section{Introduction} \label{sec:intro}
As an intuitive and efficient heuristic algorithm for sequential decision-making tasks in unknown environments, Thompson Sampling (TS) \citep{thompson1933} has been enjoying a huge success in practice and adopted in recommendation systems \citep{chapelle2011}, A/B testing \citep{graepel2010}, the online advertisement \citep{graepel2010,agarwal2013}, reinforcement learning \citep{osband2013}, etc.
Built upon online Bayesian inference framework, TS takes an action optimized to model parameters randomly drawn from the posterior distribution at each decision epoch.
This simple procedure, called \emph{posterior sampling}, finds a surprisingly proper balance between exploitation and exploration, and is proven to achieve optimality \citep{agrawal2012,russo2014}.

However, the posterior sampling procedure only considers the current level of model uncertainty, not considering the future consequences of individual actions.
This often critically affects the performance of TS, particularly when the value of exploration needs to be taken into account carefully -- for example, when there are an excessive number of arms \citep{russo2022satisficing} when the arms have different noise variances \citep{kirschner2018,min2020pg}, or when the exploration is restricted due to a budget constraint, the situation formulated as a \emph{budgeted multi-armed bandit} (MAB) \citep{ding2013,xia2015}.

In the budgeted MAB, playing an arm yields a random reward and incurs a deterministic/random cost at the same time, and no more play can be made once the playe runs out of budget.
This setting has been introduced to model online bidding optimization in sponsored search \citep{amin2012,tran2014}, and on-spot instance bidding in cloud computing \citep{agmon2013}.
The algorithms such as KUBE \citep{tran2012}, UCB-BV1/BV2 \citep{ding2013}, PD-BwK \citep{badanidiyuru2013}, i/c/m-UCB, b-Greedy \citep{xia2017}, and BTS \citep{xia2015} have been proposed and analyzed.
Budgeted Thompson Sampling (BTS), as an immediate extension of TS for budgeted MAB, is considered as a baseline algorithm to be fixed in this work.
Although it significantly outperforms the other algorithms, it still does not consider the remaining budget information when making a decision, and hence suffers from the aforementioned issue.

To overcome this shortcoming, we adopt the \textit{Information Relaxation Sampling} (IRS) framework, recently suggested by \citet{min2019thompson} for classical Bayesian $K$-armed bandit problems.
Generalizing the concept of posterior sampling, the IRS framework suggests a class of algorithms which optimize their actions to a randomly generated future scenario (not just model parameters) in a careful consideration of the belief dynamics of Bayesian learners.

Our contributions are threefold: First, by applying the IRS framework to the budgeted MAB setting, we develope a series of algorithms that can exploit the specific details of the problem instance such as budget information.
Without introducing any auxiliary parameter, they easily achieve the state-of-the-art performance.
In our numerical experiment, the improvement over BTS can be as large as 75\% in terms of reduction in regret.

Second, we obtain as byproducts a series of upper bounds on the maximal performance that can be achieved in the given problem instance.
This series of upper bounds also improve the conventional one commonly used in the definition of Bayesian regret, and turn out to be useful to see how much additional improvement can be made.

Finally, we extend IRS to random cost settings by making two levels of extensions. As a relatively simpler extension, we allow IRS policies to sample the mean cost values from their posterior distributions and then solve inner problems as if these sampled values are the ground truth, i.e., the idea of IRS is applied only to rewards but not to costs. As a more complicated extension, we can make IRS policies to sample all future cost realizations and then solve more complex inner problems that additionally consider how much the decision maker will learn about the cost distributions, i.e., the idea of IRS is applied to both rewards and costs. Our numerical experiment shows that these extensions of IRS policies indeed offer sequential improvements over BTS as expected. And it show that the more complicated extension outperforms the simple extension. 

Throughout this paper, we will focus on explaining two specific algorithms, namely, IRS.FH and IRS.V-Zero, instead of describing the general framework.

\section{Problem Formulation and Preliminaries} \label{sec:problem}
We consider a Bayesian budgeted MAB problem with $K$ arms and a resource budget $B$.
A problem instance can be specified by a tuple $\big(K,B, (c_a,\mathcal{R}_a,\Theta_a,\mathcal{P}_a,\mathcal{Y}_a,y_{a,0})_{a\in[K]} \big)$ which will be described in a greater detail below.

\paragraph{Rewards and costs.}

Let $\mathcal{A} = [K]$ be the set of arms, among which the decision maker (DM) can play one in each time period.
The stochastic reward that the DM earns from the $n^\text{th}$ pull of arm $a$ is represented with a nonnegative random variable $R_{a,n}$, and we assume that its distribution is given by $\mathcal{R}_a(\theta_a)$:
$$ R_{a,n} \sim \mathcal{R}_a(\theta_a), \quad \forall n=1,2,\ldots, $$
where $\theta_a \in \Theta_a$ is the \emph{unknown parameter} that the DM aims to learn.
Given $\theta_a$, the rewards $R_{a,1}, R_{a,2}, \ldots$ are independent.

Whenever arm $a$ is played, it also incurs a deterministic cost,\footnote{In development and analyses of our suggested algorithms, we primarily focus on the deterministic cost setting. The main ideas naturally extend to random cost setting. See \S~\ref{sec:random}.} denoted by $c_a \in \mathbb{N}$ (i.e., consumes $c_a$ units of resources deterministically).
The total amount of resources that the DM can use is limited by $B \in \mathbb{N}$, and the DM's goal is to maximize the expected total reward within this budget constraint.

\paragraph{Bayesian framework.}
In the Bayesian framework, the unknown parameter $\theta_a$ is treated as a random variable and we assume that its prior distribution is given by $\mathcal{P}_a(y_{a,0})$, i.e.,
$$ \theta_a \sim \mathcal{P}_a(y_{a,0}), $$
where the hyperparameter $y_{a,0} \in \mathcal{Y}_a$, which we call (initial) \emph{belief}, specifies the prior distribution.

As a Bayesian learner, the DM's belief about $\theta_a$ will be updated according to the Bayes' rule whenever a new reward realization from the arm $a$ is observed.
To describe the belief dynamics explicitly, we introduce a \emph{Bayesian update function} $\mathcal{U}_a: \mathcal{Y}_a \times \mathbb{R}^+ \rightarrow \mathcal{Y}_a$.
That is, after playing the arm $a$ for the first time, the belief is updated from $y_{a,0}$ to $y_{a,1} \triangleq \mathcal{U}_a(y_{a,0}, R_{a,1})$ and then the posterior distribution of $\theta_a$ can be written as $\mathcal{P}_a(y_{a,1})$.
We accordingly define $y_{a,n}$ be the belief that the DM will have after playing the arm $n$ times, i.e., $y_{a,n} \triangleq \mathcal{U}_a( y_{a,n-1}, R_{a,n} )$ for $n=1,2,\ldots$.

\paragraph{Mean reward estimates.}
We denote the unknown mean reward of arm $a$ by $\mu_a(\theta_a)$ as a real-valued function of parameter $\theta_a$:
$$ \mu_a(\theta_a) \triangleq \mathbb{E}[R_{a,n}|\theta_a ]. $$
Let us denote its $n$-sample (Bayesian) estimate by $\hat{\mu}_{a,n}(\cdot)$ as a real-valued function of first $n$ reward realizations: abbreviating $(R_{a,1}, \ldots, R_{a,n})$ as $R_{a,1:n}$, we define
\begin{equation*}
	\begin{split}
		\hat{\mu}_{a,n}( R_{a,1:n} ~; y_{a,0})
		\triangleq \mathbb{E}_{\theta_a \sim \mathcal{P}_a(y_{a,0})}[\mu_a(\theta_a)|R_{a,1:n}],
	\end{split}
\end{equation*}
which represents the expected performance of arm $a$ inferred from its first $n$ reward realizations, or equivalently, the predictive mean reward of arm $a$ that the DM would believe after playing the arm $n$ times.

These mean-reward metrics $\mu_a$ and $\hat{\mu}_{a,n}$ will be repeatedly used throughout the paper.
The reason why we define $\mu_a$ and $\hat{\mu}_{a,n}$ as functions is to clarify their dependencies on the random variables and to utilize their functional form when developing algorithms later.
To help understanding, we make the following remark.
\begin{remark}\label{rem:slln}
	By Strong Law of Large Numbers, we have
	\begin{equation*} 
		\lim_{n \rightarrow \infty} \hat{\mu}_{a,n}( R_{a,1:n} ~; y_{a,0}) = \mu_a(\theta_a), \quad \text{a.s.},
	\end{equation*}
	which says that, in terms of mean-reward estimation, knowing the parameter is equivalent to having an infinite number of observations.
	Also, for any $n$ and $k$, it holds that
$$\hat{\mu}_{a,n+k}(  R_{a,1:n+k} ~; y_{a,0}) = \hat{\mu}_{a,k}(  R_{a,n+1:n+k} ~; y_{a,n}), $$
	which says that making an inference using $n+k$ samples given an initial belief is equivalent to making an inference using the later $k$ samples after updating the belief using the former $n$ samples.
\end{remark}

\paragraph{Policy and performance.}
Let $\pi$ be the DM's policy, and $A_t$ be the arm played by $\pi$ at time $t$.
The reward that the DM earns at time $t$ can be written as
$$ r_t \triangleq R_{A_t,n_{A_t,t}}
	\quad \text{where} \quad
	n_{a,t} \triangleq \sum_{s=1}^t  \mathbf{1}\{A_s = a\}. $$
Here, $n_{a,t}$ counts the number of times that arm $a$ has been played up to time $t$.
An admissible policy $\pi$ should decide $A_t$ based only on the information revealed prior to time $t$, $(A_s, r_s)_{s=1}^{t-1}$.

Besides, playing the arm $A_t$ consumes $c_{A_t}$ units of resources.
To describe the budget constraint explicitly, we introduce a stopping time $\tau$ representing the first time that the cumulative cost exceeds the given budget, i.e., 
$$ \tau \triangleq \min\left\{ t: \sum_{s=1}^t c_{A_s} > B \right\}. $$
Only the rewards realized before time $\tau$ are counted, so the total reward collected by the DM can be written as $\sum_{t=1}^{\tau -1} r_t$.
As a trivial upper bound on $\tau$, we introduce $ T_\text{max} \triangleq \max_{a \in \mathcal{A}} \left\{ \left \lfloor B/c_a \right \rfloor+1 \right\}$.

We denote by $V(\pi)$ the expected performance of policy $\pi$ in a given MAB instance:
\begin{equation*} 
	V(\pi) \triangleq \mathbb{E}^\pi\left[ \sum_{t=1}^{\tau - 1} r_t \right] = \mathbb{E}^\pi\left[ \sum_{t=1}^{\tau - 1} \mu_{A_t}(\theta_{A_t}) \right].
\end{equation*}
Here, the expectation operator takes into account the randomness of the policy (if randomized like BTS), the reward realizations $R_{1:K,1:T_\text{max}}$, and the parameter realizations $\theta_{1:K}$.
Note that $V(\pi)$ can be alternatively represented as $\mathbb{E}^\pi\left[ \sum_{t=1}^{\tau - 1} \mu_{A_t}(\theta_{A_t}) \right]$ by the law of total expectation, since $\mathbb{E}[ r_t | A_t, \theta_{1:K} ] = \mu_{A_t}(\theta_{A_t})$.

\paragraph{Performance bound and regret.}
A quantity $W$ is said to be a \emph{performance bound} if $W \geq V(\pi)$ for any policy $\pi$.

As a performance bound commonly used in the MAB literature, $W^\text{BTS}$ is defined as\footnote{
	The naming $W^\text{BTS}$ is not common in the literature.
	The motivation for this choice is explained in \S \ref{subsec:bts}.}
\begin{equation} \label{eq:wbts}
	W^\text{BTS} \triangleq \mathbb{E}\left[B \times \max_{a \in \mathcal{A}} \frac{\mu_a(\theta_a)}{c_a}\right].
\end{equation}
This quantity represents the expected performance of the clairvoyant fractional solution: when the player knows the parameters $\theta_{1:K}$ in advance, it is optimal for him to play the arm $a^\star$ with the largest reward-to-cost ratio $\mu_a(\theta_a)/c_a$, (fractionally) $B/c_{a^\star}$ times in a row, which will yield the total reward of $\mathbb{E}\left[ \mu_{a^\star}(\theta_{a^\star}) \times B/c_{a^\star} \right]$ ($=W^\text{BTS}$) in average.
Clearly, no policy can perform better than this clairvoyant player, and therefore, $W^\text{BTS}$ is an upper bound on the maximal achievable performance for the given MAB instance.

A performance bound $W$ is said to be \emph{tighter} than the other $W'$ if $W \leq W'$.
A tighter bound provides a more precise quantification of the hardness of a particular MAB instance, and can better serve as a performance benchmark.

On the other hand, we will later utilize the  \emph{Bayesian regret} to visualize and compare the performance of policies, which is defined as
$$ \textsc{Regret}(\pi) \triangleq W^\text{BTS} - V(\pi). $$ 
The regret quantifies the suboptimality of a policy, and is non-negative since $W^\text{BTS}$ is a performance bound.
Once we have a performance bound $W$ tighter than $W^\text{BTS}$, the gap $W^\text{BTS} - W$ will provide a lower bound on the minimal achievable regret (i.e., $ \textsc{Regret}(\pi)  \geq W^\text{BTS} - W$ for any $\pi$).

\paragraph{Bayesian optimal policy.}
In the Bayesian setting, there exists a policy that achieves the maximal performance $V^\star$:
$$ V^\star \triangleq \sup_\pi V(\pi). $$
Such a \emph{Bayes-optimal policy} and its performance $V^\star$, in theory, can be obtained by solving the Bellman equation (corresponding to an MDP with a state space $\mathcal{Y}_1 \times \ldots \times \mathcal{Y}_K$ and an action space $\mathcal{A}$. See Appendix A for the detail), but they are intractable in most cases.

As motivated in the introduction, our primary goal is to improve the BTS policy in terms of performance, where the Bayes-optimal policy will be our ideal target.
Another goal is to improve the performance bound $W^\text{BTS}$ in terms of tightness, where $V^\star$ will be our ideal target.

\section{Algorithms} \label{sec:algorithm}

In this section, we propose a series of policies that improve Budgeted Thompson Sampling (BTS) toward the Bayes-optimal policy by leveraging the idea of \emph{information relaxation sampling}.
In parallel, we argue that there is a performance bound embedded in each of these policies, and accordingly, the performance bounds paired with our proposed policies also improve the performance bound paired with BTS, which is the conventional benchmark $W^\text{BTS}$.

\subsection{Budgeted Thompson Sampling} \label{subsec:bts}
As an immediate extension of Thompson Sampling to the budgeted MAB setting, BTS \citep{xia2015} utilizes the posterior sampling of the parameters.
As described in Algorithm \ref{alg:bts}, the policy $\pi^\text{BTS}$ at each time $t$ draws a random sample of the parameters from the posterior distribution (i.e., $\tilde{\theta}_a^{(t)} \sim \mathcal{P}_a(y_{a,n_{a,t-1}})$ in line 4), and plays the arm with the largest reward-to-cost ratio given the sampled parameters (i.e., $\argmax_a \mu_a(\tilde{\theta}_a^{(t)})/c_a$ in line 6).
After observing the result of the play, it updates the belief about the arm according to the Bayes' rule (line 11), and repeats this procedure until the budget is exhausted.

\begin{algorithm}[h]
\caption{BTS}
\label{alg:bts}
\textbf{Input}: $K,B, (c_a,\mathcal{R}_a,\Theta_a,\mathcal{P}_a,\mathcal{Y}_a,y_{a,0})_{a\in[K]}$ \\
\textbf{Procedure}:
\begin{algorithmic}[1]
\STATE Initialize $t \gets 1$, $B_1 \gets B$, $n_{a,0} \gets 0$ for each $a \in \mathcal{A}$
\WHILE{$B_t > 0$}
\FOR{each arm $a \in \mathcal{A}$}
\STATE Sample $\tilde{\theta}_a^{(t)} \sim \mathcal{P}_a(y_{a,n_{a,t-1}})$
\ENDFOR
\STATE $A_{t} \gets \argmax_{a \in \mathcal{A}}\{ \mu_a(\tilde{\theta}_a^{(t)}) / c_a \}$
\IF{$B_{t} < c_{A_{t}}$}
\STATE break
\ELSE
\STATE Play $A_{t}$, receive $r_{t}$, pay $c_{A_t}$ ($B_{t+1} \gets B_t - c_{A_t}$)
\STATE Update $y_{a,n_{a,t-1}+1} \gets \mathcal{U}_{a}(y_{a,n_{a,t-1}},r_{t})$, and $n_{a,t} \gets \begin{cases}n_{a,t-1}+1&\text{ for }a=A_t\\  n_{a,t-1}&\text{ for }a \ne A_t\end{cases}$
\ENDIF
\STATE $t \gets t+1$.
\ENDWHILE
\end{algorithmic}
\end{algorithm}

One can immediately relate this arm-selection rule with the performance bound $W^\text{BTS}$, defined in \eqref{eq:wbts}.
As motivated earlier, the arm $a^\star = \argmax_{a \in \mathcal{A}}\{ \mu_a(\theta_a)/c_a \}$ is the optimal one to play if the parameters are known and the fractional solution is allowed.
The policy $\pi^\text{BTS}$ mimics such a clairvoyant player's decision by replacing the unknown components $\mu_a(\theta_a)$'s with their randomly generated counterparts $\mu_a(\tilde{\theta}_a^{(t)})$'s.
Note that the randomness in this sampling procedure enforces $\pi^\text{BTS}$ to deviate from the myopic decision, resulting in explorations.

Although BTS is simple and computationally efficient ($O(K)$ computations per decision), its arm-selection rule does not incorporate the remaining budget information.
As an extreme example, if the remaining budget is so small that each arm can be play at most once, it is Bayes-optimal to make the myopic decision, i.e., $A_t \gets \argmax_{a \in \mathcal{A}}\{ \mathbb{E}_{\theta_a \sim \mathcal{P}_a(y_{a,n_{a,t-1}})}[ \mu_a(\theta_a) ] / c_a \}$.
For this reason, BTS often performs unnecessary explorations, particularly near the end of horizon, which motivates next algorithm IRS.FH.

\subsection{IRS.FH}
 
Our first proposed algorithm IRS.FH\footnote{IRS stands for Information Relaxation Sampling, and FH stands for Finite Horizon.} is very similar to BTS but additionally incorporates how many times each arm can be played in the future within the remaining budget.
While the belief updating procedure remains unchanged, IRS.FH implements a slightly different arm-selection rule, which is described in Algorithm \ref{alg:irs.fh} (lines 3--7).

\begin{algorithm}[h!]
\caption{IRS.FH}
\label{alg:irs.fh}
\textbf{Input}: $K,B, (c_a,\mathcal{R}_a,\Theta_a,\mathcal{P}_a,\mathcal{Y}_a,y_{a,0})_{a\in[K]}$ \\
\textbf{Procedure}:
\begin{algorithmic}[1] 
\STATE Initialize $t \gets 1$, $B_1 \gets B$, $n_{a,0} \gets 0$ for each $a \in \mathcal{A}$
\WHILE{$B_t > 0$}
\FOR{each arm $a \in \mathcal{A}$}
\STATE Sample $\tilde{\theta}_a^{(t)} \sim \mathcal{P}_a(y_{a,n_{a,t-1}})$
 and $\tilde{R}_{a,i}^{(t)} \sim \mathcal{R}_a(\tilde{\theta}_a^{(t)})$ for $i=1,\dots, \left\lfloor B_t/c_a \right\rfloor $
\STATE $\tilde{\hat{\mu}}_{a,\left\lfloor B_t/c_a \right\rfloor}^{(t)} \gets \hat{\mu}_{a, \left\lfloor B_t/c_a \right\rfloor }( \tilde{R}_{a,1:\left\lfloor B_t/c_a \right\rfloor}^{(t)} ; y_{a,n_{a,t-1}})$
\ENDFOR
\STATE $A_{t} \gets \argmax_{a \in \mathcal{A}} \{ \tilde{\hat{\mu}}_{a,\left\lfloor B_t/c_a \right\rfloor}^{(t)} / c_a \}$
\STATE Play $A_{t}$ and update variables (Algorithm \ref{alg:bts} lines 7--13)
\ENDWHILE
\end{algorithmic}
\end{algorithm}
\paragraph{Policy $\pi^\text{IRS.FH}$.}
More specifically, the policy $\pi^\text{IRS.FH}$ at each time samples not only the parameters $\tilde{\theta}_a^{(t)}$'s but also all future rewards $\tilde{R}_{a,i}^{(t)}$'s (line 4).
Here, $\tilde{R}_{a,i}^{(t)}$ represents the sampled reward realization associated with the future $i^\text{th}$ play of arm $a$, where $i \leq \left\lfloor B_t/c_a \right\rfloor-1$ since the arm $a$ can be updated at most $\left\lfloor B_t/c_a \right\rfloor-1$ times when the remaining budget is $B_t$.
Given these sampled future rewards, it computes the future $(\left\lfloor B_t/c_a \right\rfloor-1)$-sample mean-reward estimate $\tilde{\hat{\mu}}_{a,\left\lfloor B_t/c_a \right\rfloor-1}^{(t)}$, i.e., the belief that we would have if we allocate all remaining budget to the arm $a$ and the sampled future rewards indeed realize.
Finally, the arm with the largest reward-to-cost $\tilde{\hat{\mu}}_{a,\left\lfloor B_t/c_a \right\rfloor-1}^{(t)} / c_a$ is selected: this is almost identical to the arm-selection rule of BTS except that $\tilde{\hat{\mu}}_{a,\left\lfloor B/c_a \right\rfloor-1}^{(t)}$ is used instead of $\mu_a(\tilde{\theta}_a^{(t)})$.

In other words, $\pi^\text{IRS.FH}$ finds the best arm given a finite-number of randomly synthesized future observations.
By simulating the future belief changes using the sampled future rewards, it naturally takes into account how much we can learn in the future: when a smaller amount of budget is remaining, fewer future rewards will be sampled, and thus the future belief will less deviate from the current belief, which makes $\pi^\text{IRS.FH}$ more myopic, desirably.

Let us examine the Beta-Bernoulli case for example: when the current belief is $y_a = (\alpha_a, \beta_a)$ and the remaining budget is $B$, $\tilde{\hat{\mu}}_{a,\left\lfloor B/c_a \right\rfloor-1}$ can be expressed as
$$ \tilde{\hat{\mu}}_{a,\left\lfloor B/c_a \right\rfloor} = \frac{\alpha_a+\sum_{i=1}^{\left\lfloor B/c_a \right\rfloor-1} \tilde{R}_{a,i}}{\alpha_a+\beta_a+ \left\lfloor B/c_a \right\rfloor -1}.
$$
Note that, when $B$ is small, $\tilde{\hat{\mu}}_{a,\left\lfloor B/c_a \right\rfloor-1} \approx \frac{\alpha_a}{\alpha_a+\beta_a} = \mathbb{E}_{\theta_a \sim \text{Beta}(\alpha_a, \beta_a)}[ \mu_a(\theta_a) ]$ which leads to the myopic decision (i.e., exploitation), and when $B$ is large, $\tilde{\hat{\mu}}_{a,\left\lfloor B/c_a \right\rfloor-1} \approx \frac{1}{\left\lfloor B/c_a \right\rfloor-1} \sum_{i=1}^{\left\lfloor B/c_a \right\rfloor-1} \tilde{R}_{a,i} \approx \mu_a(\tilde{\theta}_a^t)$ which leads to the BTS's decision.
Like this, the degree of exploration is naturally adjusted depending on the amount of remaining budget, mitigating the over-exploration issue that BTS suffers from.

We also remark that IRS.FH can be computationally efficient as much as BTS.
Observe that in the above example $\sum_{i=1}^{\left\lfloor B/c_a \right\rfloor-1} \tilde{R}_{a,i}$ is distributed with $\text{Binomial}( \left\lfloor B/c_a \right\rfloor-1, \tilde{\theta}_a )$, and therefore $\tilde{\hat{\mu}}_{a,\left\lfloor B/c_a \right\rfloor-1}$ can be computed via a single random number generation without sampling $\tilde{R}_{a,i}$'s one by one.
Such a trick is applicable to more general situations where the reward distribution belongs to natural exponential family, and both IRS.FH and BTS requires $O(K)$ computations per decision.

\noindent
\paragraph{Bound $W^\text{IRS.FH}$.}
We can motivate a new performance bound $W^\text{IRS.FH}$ that is associated with $\pi^\text{IRS.FH}$.
Analogously to the way that we relate $\pi^\text{BTS}$ with $W^\text{BTS}$, we define
$$ W^\text{IRS.FH} \triangleq \mathbb{E}\left[ B \times \max_{a\in \mathcal{A}} \frac{\hat{\mu}_{a, \left\lfloor B/c_a \right\rfloor-1}(R_{a,1:\left\lfloor B/c_a \right\rfloor-1} ~; y_{a,0})}{c_a}\right]. $$
Compared to $W^\text{BTS}$, this bound implicitly postulates another type of clairvoyant player who knows $\hat{\mu}_{a, \left\lfloor B/c_a \right\rfloor}$ instead of $\mu_a(\theta_a)$.
In the task of identifying the best arm, the finite-sample mean-reward estimate $\hat{\mu}_{a, \left\lfloor B/c_a \right\rfloor}$ is less informative than the true mean-reward $\mu_a(\theta_a)$ (recall Remark \ref{rem:slln}).
Therefore, the clairvoyant player informed with $\hat{\mu}_{a, \left\lfloor B/c_a \right\rfloor-1}$ cannot perform better than the one informed with $\mu_a(\theta_a)$, which implies that $W^\text{IRS.FH}$ is tighter than $W^\text{TS}$.
We show in Theorem \ref{thm:monotonicity} that $W^\text{IRS.FH}$ is indeed a valid performance bound and improves $W^\text{BTS}$ in terms of tightness (i.e., $W^\text{BTS} \geq W^\text{IRS.FH} \geq V^\star$).

On the other hand, the value of $W^\text{IRS.FH}$ can be computed via sample averaging scheme, i.e., by repeatedly computing the term inside the expectation with respect to randomly generated $\hat{\mu}_{a, \left\lfloor B/c_a \right\rfloor}-1$'s.
This procedure can be simply implemented by reusing the code of $\pi^\text{IRS.FH}$ (lines 3--7).

\subsection{IRS.V-Zero}

We consequently propose our next algorithm, IRS.V-Zero\footnote{V-zero stands for the penalty associated with setting the prior of the information relaxation penalty discussed in $\S$\ref{subsec:genalization} to 0.}, that further improves IRS.FH by solving a more complicated optimization problem in each time period.
It takes into account not only how many times each arm can be played, but also how the belief changes over the course of future plays.

\begin{algorithm}[h]
\caption{IRS.V-Zero}
\label{alg:irs.vzero}
\textbf{Input}: $K,B, (c_a,\mathcal{R}_a,\Theta_a,\mathcal{P}_a,\mathcal{Y}_a,y_{a,0})_{a\in[K]}$ \\
\textbf{Procedure}:
\begin{algorithmic}[1] 
\STATE Initialize $t \gets 1$, $B_1 \gets B$, $n_{a,0} \gets 0$ for each $a \in \mathcal{A}$ \\
\WHILE{$B_t > 0$}
\FOR{each arm $a \in \mathcal{A}$}
\STATE Sample $\tilde{\theta}_a^{(t)} \sim \mathcal{P}_a(y_{a,n_{a,t-1}})$
 and $\tilde{R}_{a,i}^{(t)} \sim \mathcal{R}_a(\tilde{\theta}_a^{(t)})$ for $i=1,\dots,\left\lfloor B_t/c_a \right\rfloor$
\FOR{$i=1,\dots,\left\lfloor B_t/c_a \right\rfloor$}
\STATE $\tilde{\hat{\mu}}_{a,i}^{(t)} \gets \hat{\mu}_{a,i}( \tilde{R}_{a,1:i}^{(t)} ; y_{a,n_{a,t-1}})$
\ENDFOR
\ENDFOR
\STATE $$
       \text{Solve }\tilde{n}_{1:K}^\star \gets \argmax_{\tilde{n}_{1:K} \in \mathcal{N}(B_t)} \sum_{a=1}^K \sum_{i=1}^{\tilde{n}_a} \tilde{\hat{\mu}}_{a,i-1}^{(t)} \text{, where } \mathcal{N}(B_t) \triangleq \{(\tilde{n}_1,\dots,\tilde{n}_K);\sum_{a=1}^K c_a \tilde{n}_a \leq B_t\}
    $$ 
\STATE $A_{t} \gets \argmax_{a \in \mathcal{A}} \tilde{n}_a^\star$
\STATE Play $A_t$ and update variables (Algorithm \ref{alg:bts} lines 7--13)
\ENDWHILE
\end{algorithmic}
\end{algorithm}

\paragraph{Policy $\pi^\text{IRS.V-Zero}.$}
The pseudo-code is given in Algorithm \ref{alg:irs.vzero}.
The policy $\pi^\text{IRS.V-Zero}$ samples the entire future reward realizations just like $\pi^\text{IRS.FH}$ does, and computes all future finite-sample estimates $\tilde{\hat{\mu}}_{a,i}^{(t)}$ for $i=1,2,\ldots, \left\lfloor B_t/c_a \right\rfloor$ sequentially.
And then it solves a knapsack-like optimization problem (line 10) so as to determine how many times each arm should be played in the sampled future: with the nonnegative decision variables $\tilde{n}_1, \ldots, \tilde{n}_K$, it solves
\begin{equation}	\label{eq:irs.vzero}
	\text{maximize}~ \sum_{a=1}^K \sum_{i=1}^{\tilde{n}_a} \tilde{\hat{\mu}}_{a,i-1}^{(t)}
	~
	\text{subject to}~ \sum_{a=1}^K \tilde{n}_a c_a \leq B_t.
\end{equation}
Given the optimal solution $(\tilde{n}_1^\star, \ldots, \tilde{n}_K^\star) \in \mathbb{N}^K$, it actually plays the arm with the largest $\tilde{n}_a^\star$ (line 11) with an arbitrary tie-breaking rule.

The optimization problem \eqref{eq:irs.vzero} is to find the ``optimal allocation of the remaining budget across the arms''.
In its objective, the term $\tilde{\hat{\mu}}_{a,i-1}$ represents the predictive mean reward of the future $i^\text{th}$ play (predicted with the future belief right after the $(i-1)^\text{th}$ play), and the term $\sum_{i=1}^{\tilde{n}_a} \tilde{\hat{\mu}}_{a,i-1}$ represents the expected cumulative reward that can be obtained from the next $\tilde{n}_a$ plays of arm $a$.

Compared to the optimization problem that IRS.FH solves ($B \times \max_a\{\tilde{\hat{\mu}}_{a,\left\lfloor B/c_a \right\rfloor}/c_a\}$), it additionally takes into account how the belief will change over the course of the future plays, not just what the final belief will be.
This also reflects the fact that the player has to allocate $i$ plays in order to obtain the estimate $\hat{\mu}_{a,i}$.
By considering more carefully this future belief dynamics, $\pi^\text{IRS.V-Zero}$ achieves a better balance between exploitation and exploration than $\pi^\text{IRS.FH}$ does.
However, the optimization problem \eqref{eq:irs.vzero} requires $O(KBT_\text{max})$ computations to solve, which is considerably slower than IRS.FH.

\paragraph{Bound $W^\text{IRS.V-Zero}$.}
Focusing on the optimization problem \eqref{eq:irs.vzero}, we immediately obtain the following performance bound:
$$
	W^\text{IRS.V-Zero} \triangleq \mathbb{E}\left[\max_{n_{1:K}\in \mathcal{N}_B} \sum_{a=1}^K \sum_{i=1}^{n_a} \hat{\mu}_{a,i-1} \right],
$$
where $\mathcal{N}_B \triangleq \{(n_1,\dots,n_K);\sum_{a=1}^K c_a n_a \leq B \}$, and $\hat{\mu}_{a,i-1}$ hides its dependency on $R_{a,1:i-1}$ and $y_{a,0}$ for better presentation.
In Theorem \ref{thm:monotonicity}, we show that $W^\text{IRS.V-Zero}$ further improves $W^\text{IRS.FH}$.

\subsection{Generalization}\label{subsec:genalization}

Note that all of three policies, $\pi^\text{BTS}$, $\pi^\text{IRS.FH}$, and $\pi^\text{IRS.V-Zero}$, share the following structure in common: they in each time period (i) randomly generate future information via posterior sampling, (ii) optimize their decision to this randomly generated future via solving a deterministic optimization problem (referred to as \emph{inner problem}), (iii) play an arm according to the optimized decision, and update the belief according to Bayes' rule.
Their corresponding performance bounds, $W^\text{BTS}$, $W^\text{IRS.FH}$, and $W^\text{IRS.V-Zero}$, can be obtained by solving the same inner problems, not with the sampled future realizations, but with the true future realizations.

The \emph{information relaxation sampling} (IRS) framework formally generalizes this structure with the notion of information relaxation penalties.
Deferring its detailed description to Appendix A, we briefly remark that IRS unifies BTS and the Bayesian optimal policy (OPT) into a single framework, and also includes IRS.FH, IRS.V-Zero, and IRS.V-EMax as special cases that interpolate between BTS and OPT.

Each policy-bound pair is characterized by inner optimization problem: from BTS to OPT, they introduce increasingly complicated optimization problems, becoming more considerate but more computationally costly.
We indeed observe and (partly) prove that these policies achieve increasingly better performance and these performance bounds achieve increasingly better tightness.

In addition, we also implement and evaluate IRS.INDEX policy, which, strictly speaking, does not belong to IRS framework (it does not have a corresponding performance bound).
It internally utilizes IRS.V-EMax to obtain a random approximation of the Gittins index.
See Appendix A for the detail.

\section{Extension to Random Cost}\label{sec:random}
We have so far developed our framework for deterministic cost setting.
In this section, we extend IRS framework to random cost setting, in which each arm consumes a random amount of resource whenever played and this random cost is drawn from an unknown distribution that we also aim to learn.
More specifically, the stochastic cost that the DM pays for the $n^\text{th}$ pull of arm $a$ is represented with a nonnegative random variable $C_{a,n}$.
Every notation is analogously defined for costs, while we use superscript ${}^{c}$ (or ${}^{r}$) to represent the parameters/variables related to costs (or rewards, respectively):
e.g., the distribution of $C_{a,n}$ is given by $\mathcal{C}_a(\theta_a^{c})$, where $\theta_a^{c}$ is the unknown parameter for which we have a prior $\mathcal{P}_a^{c}(y_{a,0}^{c})$.

IRS algorithms can be extended to random cost in multiples ways.
We here explore two ideas --- a simple extension that uses the sampled mean cost, and a bit more complicated extension that uses the sampled future cost realizations and introduces additional penalties.

\paragraph{Simple extension}
As described in \citet{xia2015}, BTS applied to the random cost setting draws the parameters $\theta_a^{c}$'s from the posterior, and selects the arm with the largest mean-reward-to-mean-cost ratio: i.e., $\argmax_a \mu_a^r( \tilde{\theta}_a^r )/\mu_a^c( \tilde{\theta}_a^c )$.
Analogously, we motivate simple extensions of IRS policies that solve the same inner problems to the deterministic cost setting but use $\mu_a^c( \tilde{\theta}_a^c )$ instead of $c_a$.

\paragraph{Extension with additional penalties}
In the deterministic cost setting, we have motivated IRS polices by relaxing the information constraint imposed on reward realizations.
Similarly, we can consider to relax the information constraint imposed on cost realizations.
That is, we can let a policy to sample the future cost realizations in addition to the future reward realizations and solve some deterministic optimization problem with respect to this sampled future but in the presence of penalties for letting the DM exploit the future information.
A penalty function suitable for IRS.V-Zero can be designed as follows.\footnote{We extended the penalty functions not only for IRS.V-Zero but also to IRS.V-EMax and IRS.INDEX policy. The detailed procedure of two extensions is implemented in Appendix B.}

The penalty function of IRS.V-Zero is given by
$$z_t^{\text{IRS.V-Zero}}(a_{1:t},\omega) \triangleq r_t(a_{1:t},\omega) - \mathbb{E}_y[r_t(a_{1:t},\omega)|H_{t-1}(a_{1:t-1},\omega)]. $$
This penalizes the DM for knowing the future reward realizations, and similarly, we can add an extra term that penalizes the DM for knowing the future cost realizations:
\begin{equation*}
    \begin{split}
        z_t^{\text{IRS.V-Zero}}(a_{1:t},\omega)
        \triangleq r_t(a_{1:t},\omega) -& \mathbb{E}_{y^{r}}[r_t(a_{1:t},\omega)|H_{t-1}(a_{1:t-1},\omega)]\\&- \lambda \Big( c_t(a_{1:t},\omega) - \mathbb{E}_{y^{c}}[c_t(a_{1:t},\omega)|H_{t-1}(a_{1:t-1},\omega)]\Big). 
    \end{split}
\end{equation*}
Here, $\lambda \in \mathbb{R}$ supposedly captures the additional benefit that the DM can earn by knowing the actual cost realization at time $t$ instead of its expected value.
A natural choice of $\lambda$ will be the dual variable associated with the budget constraint of the inner problem that IRS.V-Zero solves, i.e., $\lambda = \max_a \mu(\theta_a^{r})/\mu(\theta_a^{c})$, the quantity reflects the additional benefit that the DM can earn when one unit of resource is additionally given.
We consider an extended version of IRS.V-Zero policy that uses its sampled value, i.e., $\tilde{\lambda} = \max_a \mu(\tilde{\theta}_a^{r})/\mu(\tilde{\theta}_a^{c})$, resulting in the following inner problem:
$$\max_{n_1,\ldots,n_K} \sum_{a=1}^K \sum_{i=1}^{n_a} \Big\{\hat{\mu}_{a,i-1}^{r} + \tilde{\lambda}(\tilde{C}_{a,i} - \hat{\mu}_{a,i-1}^{c})\Big\}
\text{ s.t. } \sum_{a=1}^K \sum_{i=1}^{n_a} \tilde{C}_{a,i} \leq B.$$

\section{Analysis} \label{sec:analysis}
We first provide a theoretical result showing that the performance bounds $W^\text{BTS}$ and $W^\text{IRS.FH}$ proposed in \S \ref{sec:algorithm} are valid upper bounds on the maximal achievable performance and incrementally tighter than the conventional benchmark. 

\begin{theorem}[Monotonicity of performance bounds] \label{thm:monotonicity}
	For any Bayesian budgeted MAB, we have
	$$ W^\text{BTS} \geq W^\text{IRS.FH} \geq W^\text{IRS.V-Zero} \geq V^\star. $$
\end{theorem}

The formal proof of Theorem \ref{thm:monotonicity} is given in Appendix C.
We briefly sketch the main idea as follows.
Recall that each of these bounds represents the maximal performance that can be achieved by a clairvoyant player who has an access to some additional information that is supposed to be unknown, and therefore, it should be greater than $V^\star$, the maximal performance of the non-clairvoyant player.
In this line of thought, the gap $W - V^\star$ can be understood as a quantity that measures how much additional benefit can be extracted by exploiting the additional information, which should decrease when less useful information is additionally given.
This explains the monotonicity $W^\text{BTS} \geq W^\text{IRS.FH} \geq W^\text{IRS.V-Zero}$, which is formally proven via Jensen's inequality.

On the other hand, the improvements in the performance bounds ($W^\text{BTS} \rightarrow W^\text{IRS.FH} \rightarrow W^\text{IRS.V-Zero}$) naturally imply the improvements in their corresponding policies ($\pi^\text{BTS} \rightarrow \pi^\text{IRS.FH} \rightarrow \pi^\text{IRS.V-Zero}$).
Recall that each of these policies mimics the behavior of the clairvoyant player using the self-generated future information, i.e., it plays an arm that would have been selected by the one who optimistically believes that the sampled future information is the ground truth.
The gap $W - V^\star$ now can be translated as a quantity that measures how overly optimistic the corresponding policy will behave.
Hence, the policy associated with a tighter performance bound is less likely to make a decision that is overly optimized to a particular realization of future information, and avoids over-explorations more effectively.

We indeed observe in all our numerical experiments that the suggested policies monotonically improve BTS in terms of performance, i.e., $V(\pi^\text{BTS}) \leq V(\pi^\text{IRS.FH}) \leq V(\pi^\text{IRS.V-Zero})$.
However, proving this monotonicity is very challenging, so we instead investigate the gaps between the performance of these policies and their corresponding performance bounds, and establish upper bounds on these gaps.

\begin{theorem}[Suboptimality gap] \label{thm:suboptimality}
	Consider a Bayesian budgeted MAB such that $\mathcal{R}_a$ is a natural exponential family distribution specified by a log-partition function $A_a(\theta_a)$ and $\mathcal{P}_a$ is given by its conjugate prior whose density function is of the form $\exp( \xi_a \theta_a - \nu_a A_a(\theta) )$.
	Suppose that all the log-partition functions are $L$-smooth, i.e., $\frac{d^2}{d\theta_a^2}A_a(\theta_a) \leq L$, $\forall \theta_a \in \Theta_a$, and $\nu_a = \nu$,  $\forall a \in \mathcal{A}$.
	Then, for any $B \geq 2 \max\{ c_1, \ldots, c_K \}$, we have
\begin{equation*}
    \begin{split}
        W^\text{BTS} - V(\pi^\text{BTS}) &\leq 2\sqrt{L}\left[\frac{1}{\sqrt{\nu}}+\sqrt{2 \log T_\text{max}} \left(\frac{K}{\sqrt{\nu}}+2\sqrt{K T_\text{max}} \right) \right],\\
        W^\text{IRS.FH} - V(\pi^\text{IRS.FH}) &\leq 2\sqrt{L}\left[\frac{1}{\sqrt{\nu}}+\sqrt{2 \log T_\text{max}} \left(\frac{K}{\sqrt{\nu}}+2\sqrt{K T_\text{max}} - \frac{1}{3} \sqrt{ \frac{T_\text{max}}{K} } \right) \right],\\
        W^\text{IRS.V-Zero} - V(\pi^\text{IRS.V-Zero}) &\leq \sqrt{L}\left[\frac{1}{\sqrt{\nu}}+\sqrt{2 \log T_\text{max}} \left(\frac{K}{\sqrt{\nu}}+2\sqrt{K T_\text{max}} - \frac{1}{3} \sqrt{ \frac{T_\text{max}}{K} } \right) \right],
    \end{split}
\end{equation*}
where $T_\text{max} \triangleq \max_{a \in \mathcal{A}} \left\{ \left \lfloor B/c_a \right \rfloor+1 \right\}$.
\end{theorem}

Theorem \ref{thm:suboptimality} considers Bayesian budgeted MABs with natural exponential family distributions, which include the Beta-Bernoulli case ($L=1/2$, $\nu=\alpha+\beta$) and the Beta-Binomial case ($L=m/2$, $\nu = (\alpha+\beta)/m$).
While all these suboptimality gaps have the same asymptotic order of $O(\sqrt{KT_\text{max} \log T_\text{max}})$, this result shows that IRS.FH and IRS.V-Zero make incremental improvements over BTS in the additional term and in the leading coefficient. The proof is given in Appendix D.

Note that our analysis aligns closely with the regret lower bound analysis and the algorithm's regret upper bound analysis typically conducted in the MAB literature. Theorem \ref{thm:monotonicity} provides a tighter lower bound $V^\star$ compared to the lower bound $W^\text{BTS}$ presented in other budgeted MAB-related studies. Theorem \ref{thm:suboptimality} highlights improvements in the suboptimality gap, distinct from the regret upper bound $W^\text{BTS} - V(\pi)$ noted in other budgeted MAB literature. The observed reduction in the suboptimality gap may be due to enhancements in $V(\pi)$, although it remains somewhat ambiguous whether these improvements are predominantly due to $W^\pi$ or $V(\pi)$. This ambiguity makes direct comparisons of $V(\pi)$ values challenging and renders the result less robust. Nevertheless, experimental evidence substantiates that notable improvements are also achieved in $V(\pi)$.

\section{Numerical Experiments} \label{sec:numerical}
We demonstrate the effectiveness of our proposed policies and performance bounds through numerical simulations.
We consider deterministic cost setting with three MAB instances\footnote{We also show that IRS algorithms are sufficiently scalable for random cost setting through numerical simulation. See Appendix E for the detail.}
 -- (a) the Beta-Bernuolli MAB with two arms, (b) the Beta-Bernuolli MAB with five arms, and (c) the Beta-Binomial MAB with six arms as a real-world example arising in the online advertisement business. In each setting, we evaluate the empirical performance of IRS policies as well as their corresponding performance bounds, and also provide a comparison with KUBE \citep{tran2012}, UCB-BV1 \citep{ding2013}, i/c/m-UCB \citep{xia2017}, and a modified version of PD-BwK \citep{badanidiyuru2013} as competing benchmarks.

\begin{figure*}
	\centering
	\begin{subfigure}[htp!]{0.3\textwidth}
		\centering
		\includegraphics[width=1.0\textwidth]{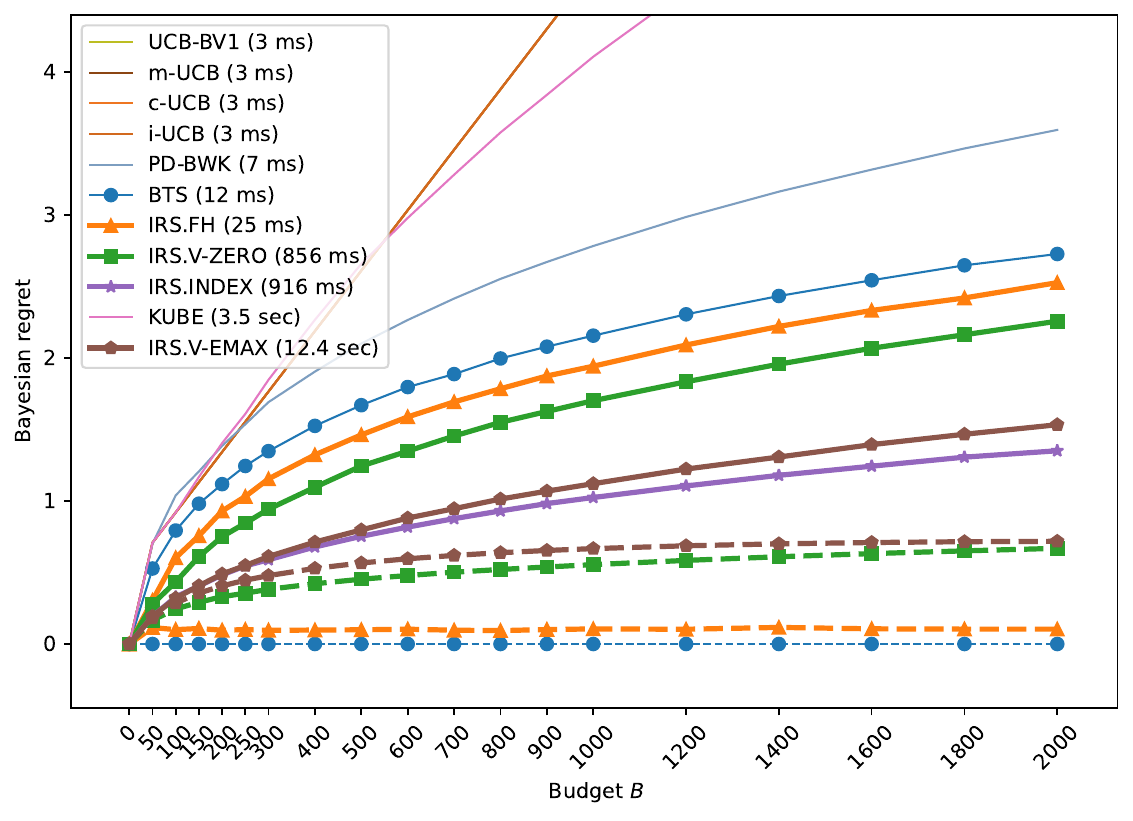}
	\end{subfigure}
	\quad
	\begin{subfigure}[htp!]{0.3\textwidth}
		\centering
		\includegraphics[width=1.0\textwidth]{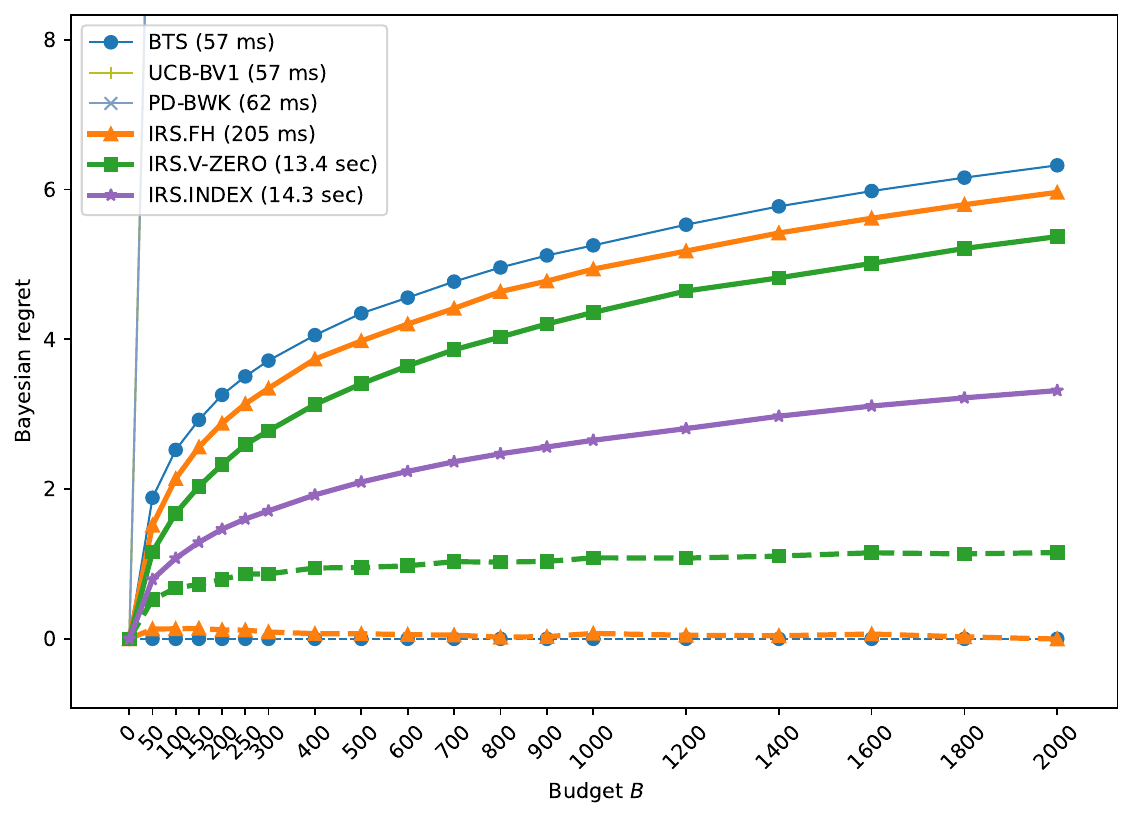}
	\end{subfigure}
	\quad
	\begin{subfigure}[htp!]{0.3\textwidth}
		\centering
		\includegraphics[width=1.0\textwidth]{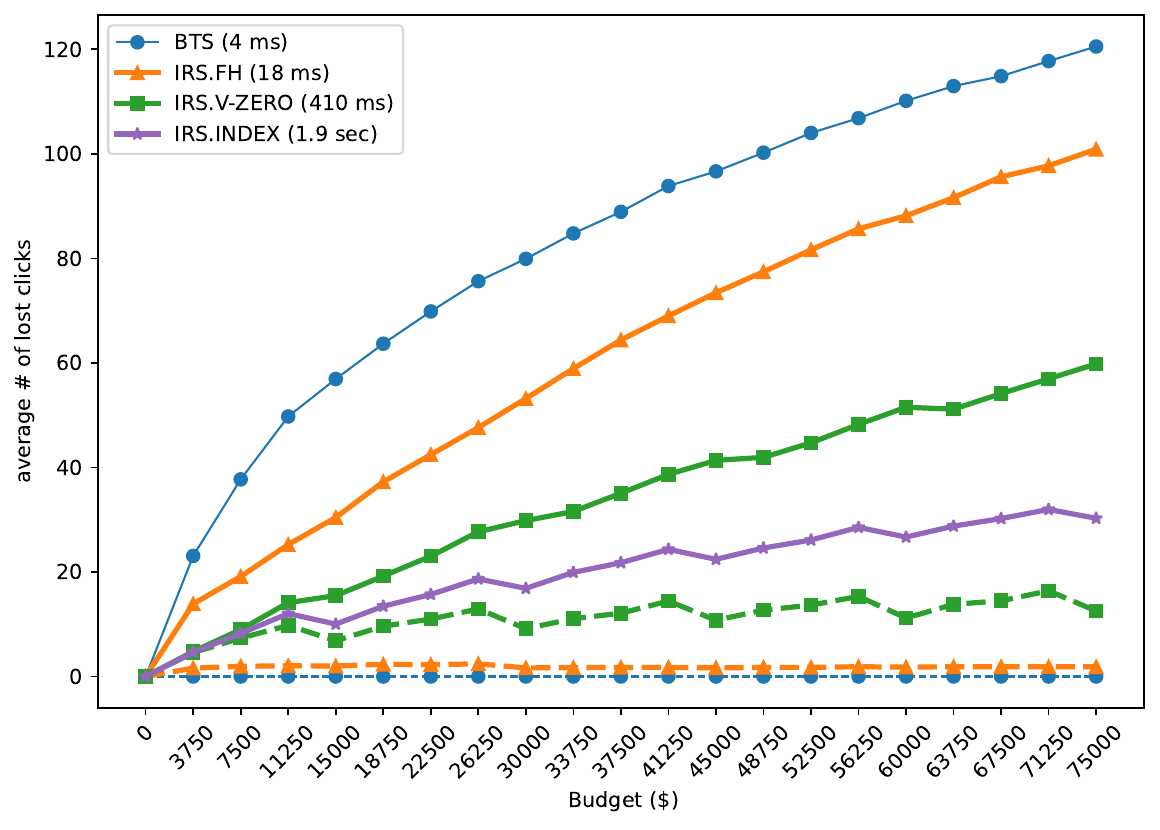}
	\end{subfigure}
	\caption{From left to right, simulation results in (a) the Beta-Bernoulli MAB with two arms, (b) the Beta-Bernoulli MAB with five arms, and (c) the Beta-Binomial MAB as a real-world example arising in online advertisement business. } 	\label{fig:simulation}
\end{figure*}

 Figure \ref{fig:simulation} visualizes the simulation results in these three settings where the $x$-axes represent the budget $B$.
The solid-line curves report the regret of the policies ($W^\text{BTS} - V(\pi)$), and the dashed-line curves report the regret lower bounds obtained with the performance bounds ($W^\text{BTS} - W$).
The run time of each policy is reported in the legend, representing the average time to complete a single run of simulation.
\paragraph{Beta-Bernoulli MABs.}
We first examine a Beta-Bernoulli MAB instance with $K=2$, $(c_1, c_2) = (10, 20)$, and $\alpha_a = \beta_a = 1, \forall a \in \mathcal{A}$, and report the result of 50,000 runs of simulation in Figure \ref{fig:simulation}(a).
When $B=2,000$, BTS outperform all competing benchmarks by a large margin, from 32\% (BTS's regret vs. PD-BwK's regret) up to 228\% (BTS's regret vs. i/c/m-UCB \& UCB-BV1's regret).
Our proposed policies even further improve BTS: IRS.FH, IRS.V-Zero, IRS.V-EMax, and IRS.INDEX policies, respectively, achieve 8\%, 18\%, 44\%, and 51\% improvement over BTS in terms of reduction in regret.
Furthermore, we can infer from the regret lower bound $W^\text{BTS} - W^\text{IRS.V-EMax}$ (brown dashed-line curve) that no policy can achieve an improvement more than 74\%, highlighting that IRS.INDEX policy is near optimal.

We next examine a Beta-Bernoulli MAB instance with $K=5$, $c_{1:5} = (2,3,10,19,20)$, and $\alpha_a = \beta_a = 1, \forall a \in \mathcal{A}$, and report the result of 20,000 runs of simulation in Figure \ref{fig:simulation}(b).
IRS.V-EMax is excluded due to its computational inefficiency.
The gaps between BTS and other benchmarks are even larger, and IRS.FH, IRS.V-Zero, and IRS.INDEX policies, respectively, achieve 6\%, 15\%, and 48\% improvement over BTS.

\paragraph{Application to online advertisement budget allocation.}
We examine a Beta-Binomial MAB instance that represents a bandit task encountered by a company who wants to optimally allocate his marketing budget across a number of ad campaigns with unknown click-through-rates (CTRs).
More specifically, the arms represent the campaigns available to this company, and playing an arm $a$ means that the company decides to spend $c_a$ dollars on the campaign $a$ on the next day which will create $m_a$ impressions.
The company's goal is to maximize the total number of clicks using a marketing budget $B$ dollars where the prior distribution of CTR can be approximated by $\text{Beta}(\alpha_a,\beta_a)$.

The problem constants $(c_a, m_a, \alpha_a, \beta_a)$'s were chosen based on iPinYou dataset \citep{liao2014}, a publicly available real-world dataset containing  logs of ad auctions, bids, impressions, clicks, and final conversions.
Imagining a region-based marketing strategy, we have empirically estimated average cost-per-impression, average number of daily impressions, and CTRs in six different regions separately, and obtained the values $(c_a, m_a, \alpha_a,\beta_a)= (\$3750, 30204, 12, 14153),(\$7200, 55965, 22, 22950),(\$15000, 120485, 25, 28968),(\$12750, 105148, 34,$ $ 44244),(\$2700, 22952, 17, 20977), (\$3300, 29847, 20, 22559)$ for $a=$ $1,\ldots,6$, respectively.
We simulate the algorithms while varying the budget $B$ from \$3,750 to \$75,000.

As shown in Figure \ref{fig:simulation}(c), IRS.FH, IRS.V-Zero, and IRS.INDEX policies, respectively, achieve 16\%, 50\%, and 75\% improvement over BTS, when the budget is \$75,000.
Converted into dollars, the improvement made by IRS.INDEX is valuable as much as \$10,000 approximately.
Given that it is impossible to reduce BTS's regret more than 89\% (as implied by $W^\text{BTS} - W^\text{IRS.V-Zero}$), we can conclude that IRS.INDEX is near-optimal.

\section{Conclusion} \label{sec:conclusion}
We have proposed a series of algorithms for budgeted MAB that improve Thompson sampling utilizing the information relaxation.
In their arm-selection procedure, they simulate Bayesian learner's belief dynamics with respect to the sampled future realizations, and by doing so they can take into account how much the decision maker can learn within the remaining budget constraint.
As a byproduct, our framework produces performance bounds that provide better quantifications of possible improvement.
While the main ideas are mostly adopted from \citet{min2019thompson}, this paper highlights that the information relaxation technique is particularly effective for budgeted bandit tasks, in which finding an optimal balance between exploration and exploitation is critical. 

Our contribution may seem obvious, but it is far from trivial. Existing literature on Budgeted MAB did not consider the use of the remaining budget information at all, and its extension in the context of the IRS framework presented its application in more realistic and appropriate settings. Unlike classic MAB problems, the termination time (the total number of pulls, denoted by stopping time $\tau$ in our proof) depends on the sequence of actions, which introduces additional challenges requiring careful theoretical analysis and complicates algorithm implementation.  

We further extend the framework to the random cost setting. The adoption of the IRS framework naturally necessitates the inclusion of cost sampling. However, a challenge arises regarding the imposition of an information relaxation penalty on cost in this context. address this challenge, we propose introducing a dual variable for the budget constraint, algorithmically simplifying it to the posterior mean reward-cost ratio. This dual variable concept holds promise for extending the imposition of additional penalties beyond budget constraints, potentially encompassing scenarios such as bandit problems with multiple constraints.

\subsubsection*{Acknowledgments}
\label{sec:ack}
This research was supported by the Asian Office of Aerospace R\&D (AOARD, Award No.: FA2386-23-1-4122).

\bibliography{main}
\bibliographystyle{rlc}

\newpage
\appendix
\section{Information Relaxation Sampling (IRS) Framework}\label{sec:IRS framework}
    
\paragraph{Additional notation.}
To describe IRS framework explicitly, we introduction some additional notation and redefine some notation defined in the main paper.
    
We define outcome (future) $\omega \in \Omega$ as the combination of parameters and all (infinitely many) future rewards, 
i.e., $\omega \triangleq (\theta,(R_{a,n})_{a \in \mathcal{A},n \in \mathbb{N}}) \sim \mathcal{I}(y)$ where $\mathcal{I}(y)$ represents the distribution for all uncertainties. 
Then we can redefine the reward $r_t$ of time $t$ defined in \S 2 as a function of $\omega$ and $a_{1:t} = (a_1, \ldots, a_t)$ as follows:
$$r_t(a_{1:t},\omega) \triangleq R_{a_t,n_t(a_{1:t},a_t)}$$
where $n_t(a_{1:t},a_t) \triangleq \sum_{s=1}^t \textbf{1}\{a_s = a\}$.
We can also redefine the stopping time $\tau$ as a function for action sequence as follows:
$$\tau(a_{1:T_{\text{max}}}) \triangleq \min\{ t ; b_t(a_{1:t}) > B\}, \quad \text{where} \quad
b_t(a_{1:t}) \triangleq \sum_{s=1}^t c_{a_s}. $$
    
Aggregating the Bayesian update function $\mathcal{U}_a$ defined for individual arms, we define $\mathcal{Y} \triangleq \mathcal{Y}_1 \times \ldots \times \mathcal{Y}_K$ and use $\mathcal{U}:\mathcal{Y}\times \mathcal{A} \times \mathbb{R} \to \mathcal{Y}$ to denote the updating function of belief vector, i.e., when a reward $r$ is observed by playing an arm $a$, so that     $\mathcal{U}(y,a,r)$ updates only the $a^\text{th}$ component of $y$.
Accordingly we define $y_{a,t}(a_{1:t},\omega)$ as
$$y_t(a_{1:t},\omega;y) \triangleq \mathcal{U}(y_{t-1}(a_{1:t-1},\omega;y),a_t,r_t(a_{1:t},\omega)), \quad \forall t \geq 1.$$
    
We define the history as $H_t(a_{1:t},\omega) \triangleq (a_1,r_1(a_1,\omega),a_2,r_2(a_{1:2},\omega),\dots,a_t,r_t(a_{1:t},\omega))$. And let $A_{1:t}^\pi$ be the action sequence up to time $t$ performed according to the DM's policy $\pi$. 
We can define the natural filtration $\mathbb{F}\triangleq (\mathcal{F}_t)_{t=0,1,\dots}$ where $\mathcal{F}_t \triangleq \sigma(H_t(A_{1:t}^\pi,\omega))$ is the $\sigma$-field generated by the history $H_t$.
A policy $\pi$ is called non-anticipating if every action $A_t^\pi$ is measurable with respect to $\mathcal{F}_{t-1}$. 
Bayesian performance of policy $\pi$ can be also redefined as,
$$V(\pi,B,y) \triangleq \mathbb{E}_y\left[ \sum_{t=1}^{\tau(A_{1:T_{\text{max}}}^\pi)-1} r_t(A_{1:t}^\pi,\omega) \right].$$

\paragraph{Bayesian optimal policy.} 
With this notation, the Bellman equation for the Budgeted MAB can be written as
\begin{equation*}
\small
\begin{split}
    Q^*(B,y,a) \triangleq 
    \begin{cases}
        \mathbb{E}_y[R_{a,1}+V^\star(B-c_a,\mathcal{U}(y,a,R_{a,1}))], & \text{if }c_a \leq B \\
        0, & \text{if } c_a > B
    \end{cases}
    , \quad
    V^\star(B,y,a) \triangleq \max_{a \in \mathcal{A}} Q^*(B, y, a).
\end{split}
\end{equation*}
    
For each time $t$, the Bayes-optimal policy (OPT) selects an arm by solving $\argmax_a Q^*(B_t,y_t,a)$ given the remaining budget $B_t$ and the current belief $y_t$. OPT achieves $V^\star$ but is difficult to solve. Therefore, \cite{min2019thompson}  introduced the information relaxation sampling (IRS).
The main idea of the IRS is to sample the future and solve the deterministic optimization problem  with penalty like other Lagrangian relaxations through the samples.
 
\subsection{Information Relaxation Performance Bound}
Here we deal more specifically with the IRS from \S 3.4. Remark that IRS relaxes the non-anticipativity constraint, allowing the DM to see the entire future but receive a information relaxation penalty. An information relaxation penalty is defined as follows.
\begin{definition}[Dual feasible penalty function, adopted from \cite{min2019thompson}]
A penalty function $z_t$ is \emph{dual feasible} if it is zero mean for any non-anticipating policy $\pi \in \Pi_\mathbb{F}$, i.e.,
$$\mathbb{E}_y \left[\sum_{t=1}^{\tau(a_{1:T_{\text{max}}})-1} z_t(A_{1:t}^\pi,\omega;B,y) \right] = 0 \quad \forall \pi \in \Pi_{\mathbb{F}}.$$
\end{definition}
Note that all penalty functions in \S 3.5 are dual feasible.

We prove that weak duality and strong duality are established even when the IRS is extended to the Budgeted MAB setting defined in $\S$2.
\begin{theorem}[Weak duality and strong duality, adopted from \cite{min2019thompson}] \label{thm:duality}
If the penalty function $z_t$ is dual feasible, $W^z$ is an upper bound on the optimal value $V^\star$:
$$\text{(Weak duality)  } W^z(B,y) \geq V^\star(B,y).$$
There exists a dual feasible penalty function denoted by $z_t^{\text{ideal}}$ such that
$$\text{(Strong duality)  } W^{\text{ideal}}(B,y) = V^\star(B,y), $$
where the ideal penalty function $z_t^{\text{ideal}}$ has the following functional form:
\begin{equation*}
\begin{split}
z_t^{\text{ideal}}(a_{1:t},\omega;B,y) \triangleq& r_t(a_{1:t},\omega) - \mathbb{E}_y[r_t(a_{1:t},\omega)|H_{t-1}(a_{1:t-1},\omega)]\\
&+ V^\star(B-b_t(a_{1:t}),y_{t}(a_{1:t},\omega;y))- \mathbb{E}_y[V^\star(B-b_t(a_{1:t}),y_{t}(a_{1:t},\omega;y))|H_{t-1}(a_{1:t-1},\omega)].
\end{split}
\end{equation*}
\end{theorem}
\emph{(Proof of weak duality)}
Let $\oplus$ be operator that adds elements to the vector, i.e., $a_{1:t} = a_{1:t-1} \oplus a_t$.
And define the filtration for the perfect relaxations $\mathcal{G}_t \triangleq \mathcal{F}_t \cup \sigma(\omega)$. 
Then we consider a relaxed policy space $\Pi_{\mathbb{G}}\triangleq\{\pi:A_t^{\pi} \text{ is } \mathcal{G}_{t-1}\text{-measurable}, \forall t \}$ and we have,
\begin{equation*}   
\begin{split}
V^\star(B,y) \triangleq& \sup_{\pi \in \Pi_{\mathbb{F}}} \mathbb{E}\Bigl[\sum_{t=1}^{\tau(A_{1:T_{\text{max}}}^{\pi})-1} r_t(A_{1:t}^{\pi},\omega) \Bigl] 
= \sup_{\pi \in \Pi_{\mathbb{F}}} \mathbb{E}\Bigl[\sum_{t=1}^{\tau(A_{1:T_{\text{max}}}^{\pi})-1}  r_t(A_{1:t}^{\pi},\omega) - z_t(A_{1:t}^{\pi},\omega) \Bigl]
\\ \leq& \sup_{\pi \in \Pi_{\mathbb{G}}} \mathbb{E}\Bigl[\sum_{t=1}^{\tau(A_{1:T_{\text{max}}}^{\pi})-1}  r_t(A_{1:t}^{\pi},\omega) - z_t(A_{1:t}^{\pi},\omega) \Bigl]
\\=& \mathbb{E}\Bigl[\max_{a_{1:T_{\text{max}}}\in \mathcal{A}^{T_{\text{max}}}}\sum_{t=1}^{\tau(a_{1:T_{\text{max}}})-1} r_t(a_{1:t},\omega) - z_t(a_{1:t},\omega) \Bigl]
= W^z(B,y),
\end{split}
\end{equation*}
which concludes the proof. $\blacksquare$
\\
    
\noindent
\emph{(Proof of Strong duality)}
Without loss of generality, we set $V^\star(B, y) = 0$ if $B < 0$ and $Q^\star(B,y,a) = 0$ if $B < c_a$.
Fix $B$ and $y$. Let the value function and Q-value function of the inner problem be $V_{t+1}^{\text{in}}(a_{1:t-1},\omega)$ and $Q_t^{\text{in}}(a_{1:t-1},a,\omega)$, respectively. Then with $V_t^{\text{in}}(a_{1:t}, \omega) = 0$ for $b_t(a_{1:t}) > B$, we can write the Bellman equation for inner problem as
\begin{equation*}
\small
\begin{split}
Q_t^{\text{in}}(a_{1:t-1},a,\omega) =
\left\{ \begin{array}{ll}
r_t(a_{1:t-1} \oplus a, \omega) - z_t^{\text{ideal}}(a_{1:t-1}\oplus a, \omega) + V_{t+1}^{\text{in}}(a_{1:t-1}\oplus a, \omega), & \text{if } b_t(a_{1:t-1}\oplus a) \leq B \\
0 & \text{if } b_t(a_{1:t-1}\oplus a) > B
\end{array} \right.
		,
\end{split}
\end{equation*}
\begin{equation*}
\begin{split}
V_t^{\text{in}}(a_{1:t-1},\omega) = \max_{a \in \mathcal{A}}Q_t^{\text{in}}(a_{1:t-1},a,\omega).
\end{split}
\end{equation*}

We argue by induction to show that for all $a_{1:t-1}\in \mathcal{A}^{t-1}, a \in \mathcal{A}, t \in \{1, \ldots, T_\text{max}+1\}$, 
$$V_t^{\text{in}}(a_{1:t-1},\omega) = V^\star(B-b_{t-1}(a_{1:t-1}),y_{t-1}(a_{1:t-1},\omega)), $$
$$Q_t^{\text{in}}(a_{1:t-1},a,\omega) = Q^\star(B-b_{t-1}(a_{1:t-1}),y_{t-1}(a_{1:t-1},\omega),a). $$

When $t=T_\text{max}+1$,
\begin{equation*}
\begin{split}
V_t^{\text{in}}(a_{1:t-1},\omega) = 0= V^\star(B-b_{t-1}(a_{1:t-1}), y_{t-1}(a_{1:t-1},\omega))  
\end{split}
\end{equation*}
since $B < b_{T_\text{max}+1}(a_{1:T_\text{max}+1})$ for any $a_{1:T_\text{max}+1}$.
    
Assume that the claim holds for $t+1$: $V_{t+1}^{\text{in}}(a_{1:t},\omega) = V^\star(B-b_{t}(a_{1:t}),y_{t}(a_{1:t},\omega))$.
Then, for any $a_{1:t-1} \in \mathcal{A}^{t-1}$ and $a \in \mathcal{A}$, if $b_t(a_{1:t-1} \oplus a ) \leq B$,
\begin{equation*}
\begin{split}
Q_t^{in}(a_{1:t-1},a,\omega) =& r_t(a_{1:t-1} \oplus a, \omega) - z_t^{ideal}(a_{1:t-1}\oplus a, \omega) + V_{t+1}^{in}(a_{1:t-1}\oplus a, \omega)\\
            =& \mathbb{E}_y[r_t(a_{1:t-1}\oplus a,\omega)+ V^*(B-b_t(a_{1:t-1}\oplus a),y_{t}(a_{1:t-1}\oplus a,\omega;y))|H_{t-1}(a_{1:t-1},\omega)]\\
            &\underbrace{-V^*(B-b_t(a_{1:t-1}\oplus a),y_{t}(a_{1:t-1}\oplus a,\omega;y)) +  V_{t+1}^{in}(a_{1:t-1}\oplus a, \omega)}_{=0}\\
            &= \mathbb{E}_y[r_t(a_{1:t-1}\oplus a,\omega)+ V^*(B-b_t(a_{1:t-1}\oplus a),y_{t}(a_{1:t-1}\oplus a,\omega;y))|H_{t-1}(a_{1:t-1},\omega)]\\
            &= \mathbb{E}_{y_{t-1}(a_{1:t-1},\omega)}[R_a + V^*(B-b_{t-1}(a_{1:t-1})-c_a,\mathcal{U}(y_{t-1}(a_{1:t-1},\omega),a,R_a)]\\
            &= Q^*(B-b_{t-1}(a_{1:t-1}),y_{t-1}(a_{1:t-1},\omega),a),
\end{split}
\end{equation*}
and if $b_t(a_{1:t-1} \oplus a ) > B$, $Q_t^{in}(a_{1:t-1},a,\omega)  = 0 = Q^*(B-b_{t-1}(a_{1:t-1}),y_{t-1}(a_{1:t-1},\omega),a)$.

Therefore,
\begin{equation*}
\begin{split}
            V_t^{\text{in}}(a_{1:t-1},\omega) =& \max_{a \in \mathcal{A}}\{Q_t^{\text{in}}(a_{1:t-1},a,\omega)\}\\
            =& \max_{a \in \mathcal{A}}\{Q^*(B-b_{t-1}(a_{1:t-1})-c_a,y_{t-1}(a_{1:t-1},\omega),a)\}\\
            =& V^\star(B-b_{t}(a_{1:t}),y_{t-1}(a_{1:t-1},\omega)).
\end{split}
\end{equation*}
    
Therefore the claim holds for all $t= 1,\cdots,T_\text{max}$. In particular for $t=1$, we have.
\[V_1^{\text{in}}(\varnothing,\omega) =V^\star(B,y), \quad Q_1^{\text{in}}(\varnothing,a,\omega) =Q^*(B,y,a), \quad \forall \omega. \hfill \blacksquare\]

\subsection{Generic IRS policy }

The \emph{information relaxation sampling} (IRS) framework formally generalizes this structure with the notion of information relaxation penalties.
The IRS relaxes the non-anticipativity constraint, allowing the DM to see the entire future but receive a penalty, so called \textit{information relaxation penalty}.
Let $z_t(a_{1:t},\omega;B,y)$ be the penalty function received at time $t$, where $\omega$ is the given true future realizations.
 Then, the deterministic optimization problem (inner problem) solved by the DM who knows the entire future $\omega$ is
$$ \max_{a_{1:T_{\text{max}}} \in \mathcal{A}^{T_{\text{max}}}}\left\{ \sum_{t=1}^{\tau(a_{1:T_{\text{max}}})-1} r_t(a_{1:t},\omega) - z_t(a_{1:t},\omega;B,y) \right\}. $$ 

An IRS performance bound induced by the penalty function $z_t(\cdot)$ is defined as
$$ W^z(B,y) =\mathbb{E}_y\left[\max_{a_{1:T_{\text{max}}}\in \mathcal{A}^{T_{\text{max}}}}\left\{\sum_{t=1}^{\tau(a_{1:T_{\text{max}}})-1} r_t(a_{1:t},\omega) - z_t(a_{1:t},\omega;B,y) \right\} \right].$$

And IRS policy that solves the inner problem with respect to the randomly generated outcome $\tilde{\omega}$ sampled using posterior sampling instead of true outcome $\omega$. Given a penalty function $z_t$, we can specify the non-anticipating policy $\pi^z$ as in Algorithm \ref{alg:irs}. 
Considering the recursive structure of the Bayesian Budgeted MAB, it repeatedly decides which arm to play based only on the current belief $y_{t-1}$ and the remaining budget $B_t$. That is, it plays the action that would have been selected in the first period in an MAB instance specified by $B_t$ and $y_{t-1}$. 

\begin{algorithm}[h!]
\caption{Generic IRS policy}
\label{alg:irs}
\textbf{Input}: MAB instance $\big( K,B, (c_a,\mathcal{R}_a,\Theta_a,\mathcal{P}_a,\mathcal{Y}_a,y_{a,0})_{a\in[K]} \big)$, penalty function $z_t(\cdot)$ \\
\textbf{Procedure}:
\begin{algorithmic}[1] 
\STATE Initialize $t \gets 1$, $B_1 \gets B$, $y_0 \gets (y_{a,0})_{a \in \mathcal{A}}$ \\
\WHILE{$B_t > 0$}
\STATE Sample $\tilde{\omega} \sim \mathcal{I}(y_{t-1})$:  i.e., sample $\tilde{\theta}_a^{(t)} \sim \mathcal{P}_a([y_{t-1}]_a)$\\ \ \ \ and $\tilde{R}_{a,i}^{(t)} \sim \mathcal{R}_a(\tilde{\theta}_a^{(t)})$ for $i=1,\dots,\left\lfloor B_t/c_a \right\rfloor$ for each arm $a \in \mathcal{A}$
\STATE $\tilde{a}_{1:T_{\text{max}}}^\star \gets \max_{a_{1:T_{\text{max}}} \in \mathcal{A}^{T_{\text{max}}}}\Bigl\{ \sum_{t=1}^{\tau(a_{1:T_{\text{max}}})-1} r_t(a_{1:t},\tilde{\omega}) - z_t(a_{1:t},\tilde{\omega};B,y) \Bigl\}$
\STATE $A_{t} \gets \tilde{a}_1^\star$
\IF{$B_{t} < c_{A_{t}}$}
\STATE break
\ELSE
\STATE Play $A_{t}$, receive $r_{t}$, pay $c_{A_t}$ ($B_{t+1} \gets B_t - c_{A_t}$)
\STATE Update $y_{a,n_{a,t-1}+1} \gets \mathcal{U}_{a}(y_{a,n_{a,t-1}},r_{t})$, and $n_{a,t} \gets \begin{cases}n_{a,t-1}+1&\text{ for }a=A_t\\  n_{a,t-1}&\text{ for }a \ne A_t\end{cases}$
\ENDIF
\STATE $t \gets t+1$ 
\ENDWHILE
\end{algorithmic}
\end{algorithm}
    
We remark that IRS unifies BTS and the Bayesian optimal policy (OPT) into a single framework, and also includes IRS.FH, IRS.V-Zero, and IRS.V-EMax as special cases that interpolate between BTS and OPT. More specifically, in the IRS framework, we can get the performance bound $W^z$ and the policy $\pi^z$ by choosing the dual feasible penalty $z_t$.  Following \cite{min2019thompson}, we consider a specific family of penalty functions:
\begin{equation*}\small  \begin{split}
        z_t^{\text{BTS.I}}(a_{1:t},\omega) &\triangleq  r_t(a_{1:t},\omega) - \mathbb{E}_y[r_t(a_{1:t},\omega)|\theta], \\
        z_t^{\text{IRS.FH.I}}(a_{1:t},\omega) &\triangleq r_t(a_{1:t},\omega) - \mathbb{E}_y[r_t(a_{1:t},\omega)|(\hat{\mu}_{a, \left\lfloor B/c_a\right\rfloor-1}(\omega))_{a \in \mathcal{A}}], \\
        z_t^{\text{IRS.V-Zero}}(a_{1:t},\omega) &\triangleq r_t(a_{1:t},\omega) - \mathbb{E}_y[r_t(a_{1:t},\omega)|H_{t-1}(a_{1:t-1},\omega)], \\ 
        z_t^{\text{IRS.V-EMax}}(a_{1:t},\omega) &\triangleq r_t(a_{1:t},\omega) - \mathbb{E}_y[r_t(a_{1:t},\omega)|H_{t-1}(a_{1:t-1},\omega), ]\\ 
        &+ W^\text{BTS}(B-b_t(a_{1:t}),y_{t}(a_{1:t},\omega;y)) - \mathbb{E}_y[W^\text{BTS}(B-b_t(a_{1:t}),y_{t}(a_{1:t},\omega;y))|H_{t-1}(a_{1:t-1},\omega)],\\
	z_t^{\text{ideal}}(a_{1:t},\omega;B,y) &\triangleq r_t(a_{1:t},\omega) - \mathbb{E}_y[r_t(a_{1:t},\omega)|H_{t-1}(a_{1:t-1},\omega)]\\ 
	&+ V^\star(B-b_t(a_{1:t}),y_{t}(a_{1:t},\omega;y)) - \mathbb{E}_y[V^\star(B-b_t(a_{1:t}),y_{t}(a_{1:t},\omega;y))|H_{t-1}(a_{1:t-1},\omega)],
\end{split}
\end{equation*}
each of which will be discussed in detail in Appendix A but we briefly explain about $z_t^{\text{IRS.V-EMax}}$ and $z_t^{\text{ideal}}$. Let $W^z$ is an upper bound on the optimal value $V^\star$. Then, $z_t^{\text{ideal}}$ is a penalty function that makes $W^z = V^\star$, and $z_t^{\text{IRS.V-EMax}}$ is a penalty function that replaces $V^\star$ of $z_t^{\text{ideal}}$, which is difficult to find, with $W^{\text{BTS}}$. 

\subsection{BTS Revisited}
Before proceed, we define
$$ \mathcal{N}^I(B) \triangleq \left\{n_{1:K} \in \mathbb{N}_{0}^K; \sum_{a=1}^K c_a n_a \leq B \right\}
, \quad \mathcal{N}^R(B) \triangleq \left\{n_{1:K} \in \mathbb{R}_{0}^K; \sum_{a=1}^K c_a n_a \leq B \right\}, $$
where $\mathbb{N}_0$ denotes the set of non-negative integers, and $\mathbb{R}_0$ denotes the non-negative real numbers.

We first consider the penalty function $z_t^{\text{BTS.I}}(\cdot)$:
\begin{equation*}
\begin{split}
z_t^{\text{BTS.I}}(a_{1:t},\omega) \triangleq r_t(a_{1:t},\omega) -  \mathbb{E}_y[r_t(a_{1:t},\omega)|\theta] = r_t(a_{1:t},\omega) - \mu_{a_t}(\theta_{a_t}).
\end{split}
\end{equation*}
    
The associated inner problem becomes
\begin{equation*}
\begin{split}
\max_{a_{1:T_{\text{max}}} \in \mathcal{A}^{T_{\text{max}}}}\Biggl\{ \sum_{t=1}^{\tau(a_{1:T_{\text{max}}})-1} r_t(a_{1:t},\omega) - z_t^{\text{BTS.I}}(a_{1:t},\omega) \Biggr\} 
&= \max_{a_{1:T_{\text{max}}} \in \mathcal{A}^{T_{\text{max}}}}\Biggl\{ \sum_{t=1}^{\tau(a_{1:T_{\text{max}}})-1} \mu_{a_t}(\theta_{a_t}) \Biggr\} \\
&= \max_{n_{1:K} \in \mathcal{N}^I(B)} \Biggl\{ \sum_{a=1}^K \mu_a(\theta_a) \cdot n_a \Biggr\}.
\end{split}
\end{equation*}
By applying the fractional relaxation (i.e., by allowing this optimization problem to admit fractional solutions), since $\mathcal{N}^I(B) \subset \mathcal{N}^R(B)$,
$$ 
\max_{n_{1:K} \in \mathcal{N}^I(B)} \Biggl\{ \sum_{a=1}^K \mu_a(\theta_a) \cdot n_a \Biggr\}
~\leq~ \max_{n_{1:K} \in \mathcal{N}^R(B)} \Biggl\{ \sum_{a=1}^K \mu_a(\theta_a) \cdot n_a \Biggr\}
= B \times \max_{a \in \mathcal{A}} \frac{\mu_a(\theta_a)}{c_a}. $$
Consequently, the performance bound is
$$
W^{\text{BTS.I}}(B,y) 
:= \mathbb{E}_y\Biggl[\max_{n_{1:K} \in \mathcal{N}^I(B)} \Biggl\{ \sum_{a=1}^K \mu_a(\theta_a) \cdot n_a \Biggr\}\Biggr]
~\leq~ \mathbb{E}_y\Biggl[  B \times \max_{a \in \mathcal{A}} \frac{\mu_a(\theta_a)}{c_a} \Biggr]
=: W^\text{BTS}(B,y),
$$
where $W^\text{BTS}(B,y)$ is the conventional benchmark that is the performance bound associated with the BTS policy.
	
Since the penalty function is dual feasible, by weak duality,
$$  W^{\text{BTS.I}}(B,y) \geq V^\star(B,y), $$
which implies that $W^\text{BTS}(B,y)$ is also a valid performance bound.
   
In other words, the policy $\pi^\text{BTS}$ and bound $W^\text{BTS}$ are the ones induced by the penalty function $z_t^{\text{BTS.I}}(\cdot)$ together with the \emph{fractional relaxation} of the inner problem.
   
\subsection{IRS.FH Revisited}
We introduce the penalty function $z_t^{\text{IRS.FH.I}}(\cdot) $ defined as
\begin{equation*}
\begin{split}
z_t^{\text{IRS.FH.I}}(a_{1:t},\omega) 
\triangleq r_t(a_{1:t},\omega) - \mathbb{E}_y[r_t(a_{1:t},\omega)|(\hat{\mu}_{\left\lfloor B/c_a\right\rfloor-1}(\omega))_{a \in \mathcal{A}}]
= r_t(a_{1:t},\omega) - \hat{\mu}_{a_t,\left \lfloor B/c_{a_t} \right \rfloor -1}(\omega) .
\end{split}
\end{equation*}
The associated inner problem satisfies
\begin{equation*}
\begin{split}
            \max_{a_{1:T_{\text{max}}} \in \mathcal{A}^{T_{\text{max}}}}\Bigl\{ \sum_{t=1}^{\tau(a_{1:T_{\text{max}}})-1} r_t(a_{1:t},\omega) - z_t^{\text{IRS.FH.I}}(a_{1:t},\omega)  \Bigl\} 
            &= \max_{a_{1:T_{\text{max}}} \in \mathcal{A}^{T_{\text{max}}}}\Bigl\{ \sum_{t=1}^{\tau(a_{1:T_{\text{max}}})-1}  \hat{\mu}_{a_t,\left \lfloor B/c_{a_t} \right \rfloor -1}(\omega) \Bigl\}
            \\&= \max_{n_{1:K} \in \mathcal{N}^I(B)} \Bigl\{ \sum_{a=1}^K \hat{\mu}_{a,\left \lfloor B/c_a \right \rfloor -1}(\omega) \cdot n_a\Bigl\}
            \\&\leq \max_{n_{1:K} \in \mathcal{N}^R(B)} \Bigl\{ \sum_{a=1}^K \hat{\mu}_{a,\left \lfloor B/c_a \right \rfloor -1}(\omega) \cdot n_a\Bigl\}
            \\&= B \times \max_{a \in \mathcal{A}} \frac{\hat{\mu}_{a,\left \lfloor B/c_a \right \rfloor -1}(\omega)}{c_a},
\end{split}
\end{equation*}
which says that the inner problem solved by the IRS.FH algorithm is also fractional relaxation of the inner problem induced by the penalty function $z_t^{\text{IRS.FH.I}}(\cdot)$
    
Since this penalty function is dual feasible,
\begin{equation*}
\begin{split}
  	V^\star(B,y) \leq W^{\text{IRS.FH.I}}(B,y) &:= \mathbb{E}\left[ \max_{n_{1:K} \in \mathcal{N}^I(B)} \Bigl\{ \sum_{a=1}^K \hat{\mu}_{a,\left \lfloor B/c_a \right \rfloor -1}(\omega) \cdot n_a\Bigl\} \right]\\
		&\leq \mathbb{E}\left[ B \times \max_{a \in \mathcal{A}} \frac{\hat{\mu}_{a,\left \lfloor B/c_a \right \rfloor -1}(\omega)}{c_a} \right]
		=: W^\text{IRS.FH}(B,y).
\end{split}
\end{equation*}
  
\subsection{IRS.V-Zero Revisited}
We consider the following penalty function
$$z_t^{\text{IRS.V-Zero}}(a_{1:t},\omega) \triangleq r_t(a_{1:t},\omega) - \mathbb{E}_y[r_t(a_{1:t},\omega)|H_{t-1}(a_{1:t-1},\omega)].$$
It follows that
 $$r_t(a_{1:t},\omega) - z_t^{\text{IRS.V-Zero}}(a_{1:t},\omega) = \mathbb{E}_y[r_t(a_{1:t},\omega)|H_{t-1}(a_{1:t-1},\omega)]
=  \hat{\mu}_{a_t,n_{t-1}(a_{1:t-1},a_t)}(\omega;y_a). $$
With $S_{a,n}(\omega) \triangleq \sum_{i=1}^n \hat{\mu}_{a,i-1}(\omega)$ representing the cumulative payoff from the first $n$ pulls of an arm $a$, the inner problem can be written as
$$\max_{a_{1:T_{\text{max}}} \in \mathcal{A}^{T_{\text{max}}}} \Bigl\{ \sum_{t=1}^{\tau(a_{1:T_{\text{max}}})-1} \hat{\mu}_{a_t,n_{t-1}(a_{1:t-1},a_t)} \Bigl\}  = \max_{n_{1:K} \in \mathcal{N}^I(B)} \Bigl\{ \sum_{a=1}^K S_{a,n_a}\Bigl\},$$
which is the exactly same inner problem that IRS.V-Zero solves.
    
\subsection{IRS.V-EMax}
Recall the functional form of $z^{\text{ideal}}$ in Theorem \ref{thm:duality}:
\begin{equation*}
\begin{split}
z_t^{\text{ideal}}(a_{1:t},\omega;B,y) &\triangleq r_t(a_{1:t},\omega) - \mathbb{E}_y[r_t(a_{1:t},\omega)|H_{t-1}(a_{1:t-1},\omega)]\\ 
&+ V^\star(B-b_t(a_{1:t}),y_{t}(a_{1:t},\omega;y)) - \mathbb{E}_y[V^\star(B-b_t(a_{1:t}),y_{t}(a_{1:t},\omega;y))|H_{t-1}(a_{1:t-1},\omega)]
\end{split}
\end{equation*}
There is a penalty not only for knowing future rewards, but also for knowing how future beliefs evolve according to action sequence.
Explicitly, the penalty for rewards is $r_t(a_{1:t},\omega) - \mathbb{E}_y[r_t(a_{1:t},\omega)|H_{t-1}(a_{1:t-1},\omega)]$, and we defined it as $z_t^{\text{IRS.V-Zero}}$.
    
$z_t^{\text{IRS.V-EMax}}$ is a penalty that even adds a penalty for knowing how future beliefs evolve according to action sequence. However, $V^\star$ is difficult to calculate because it has to solve the bellman equation, so IRS.V-EMax replaces $V^\star$ with $W^{\text{BTS}}$.
Also note that
\begin{equation*}
\begin{split}
            \mathbb{E}[W^\text{BTS}(B-b_{t}(a_{1:t}),y_{t}(a_{1:t},\omega;y))|H_{t-1}] =& (B-b_{t}(a_{1:t})) \times \mathbb{E}_y[\max_a \frac{\mu_a(\theta_a)}{c_a}|H_{t-1}] \\
            =& (B-b_{t}(a_{1:t})) \times E_{y_{t-1}}[\max_a \frac{\mu_a(\theta_a)}{c_a}] \\
            =& W^\text{BTS}(B-b_{t}(a_{1:t}),y_{t-1}(a_{1:t-1},\omega;y))
\end{split}
\end{equation*}
So we can rewrite $z_t^{\text{IRS.V-EMax}}$ as
\begin{equation*}
\begin{split}
z_t^{\text{IRS.V-EMax}}\triangleq& r_t(a_{1:t},\omega) - \mathbb{E}_y[r_t(a_{1:t},\omega)|H_{t-1}(a_{1:t-1},\omega)]\\ 
&+ W^\text{BTS}(B-b_t(a_{1:t}),y_{t}(a_{1:t},\omega;y))- \mathbb{E}_y[W^\text{BTS}(B-b_t(a_{1:t}),y_{t}(a_{1:t},\omega;y))|H_{t-1}(a_{1:t-1},\omega)]\\
=&z_t^{\text{IRS.V-Zero}} + W^\text{BTS}(B-b_t(a_{1:t}),y_{t}(a_{1:t},\omega;y))-W^\text{BTS}(B-b_{t}(a_{1:t}),y_{t-1}(a_{1:t-1},\omega;y))
\end{split}
\end{equation*}
    
Given $\omega$, then belief $y_{t}(a_{1:t},\omega)$ depends on the action sequence. Therefore, IRS.V-EMax solves the inner problem of finding the optimal action sequence rather than finding the optimal allocation of the number of pulls.
    
\paragraph{How to solve the inner problem.}
For better presentation, we temporarily refine some notation as a function of the past number of plays, not the past action sequence.
Let $y(n_{1:K},\omega)$ be the belief as a function of pull counts and $b(n_{1:K})\triangleq \sum_{a=1}^K c_a \times n_a$ be the consumed budget as a function of pull counts.
Then we can write the payoff from arm $a$ given the past number of plays $n_{1:K}$ in the inner problem solved by IRS.V-EMax as
\begin{equation*}
r^z(n_{1:K},a,\omega) \triangleq 
\left\{ \begin{array}{ll} 
            	
\begin{split}
&\hat{\mu}_{a,n_a}(\omega)-  W^\text{BTS}(B-b(n_{1:K}+e_a),y(n_{1:K}+e_a,\omega))\\&
\quad	+W^\text{BTS}(B-b(n_{1:K}+e_a),y(n_{1:K},\omega)) 
\end{split} & \text{if } b(n_{1:K}+e_a) \leq B \\
0 & \text{if } b(n_{1:K}+e_a) > B.
\end{array} \right.
\end{equation*}
where $e_a \in \mathbb{N}_0^K$  is a basis vector such that the $a^{th}$ component is one and the others are zero.

And we denote the subproblem of the inner problem of IRS.V-EMax given $n_{1:K}$ as follows:
\begin{equation*}
\begin{split}
M(n_{1:K},\omega) \triangleq \max_{a_{1:t}\in \mathcal{A}^t}\Bigl\{\sum_{s=1}^{t} r_s(a_{1:s},\omega)- z_s^{\text{IRS.V-EMax}} ;\sum_{s=1}^t \textbf{1}\{a_s = a\} = n_a, \forall a\Bigl\},
\end{split}
\end{equation*}
where $t = \sum_{a=1}^K n_a$.
Then maximal value of $M(n_{1:K},\omega)$ must satisfy the following Bellman equation:
\begin{equation*}
\begin{split}
M(n_{1:K},\omega) = \max_{a \in \mathcal{A}} \Bigl\{ M(n_{1:K} - e_a,\omega) + r^z(n_{1:K}-e_a, a, \omega) \Bigl\}.
\end{split}
\end{equation*}
Therefore, the maximal value of the inner problem can be obtained by finding $n_{1:K}^\star$ that maximizes $M(n_{1:K},\omega)$, i.e.,
$$\max_{n_{1:K} \in \mathcal{N}^I(B)} \{M(n_{1:K},\omega)\}.$$

\begin{algorithm}[htp!]
\caption{IRS.V-EMax}
\label{alg:irs.vemax}
\textbf{Input}: $K,B, (c_a,\mathcal{R}_a,\Theta_a,\mathcal{P}_a,\mathcal{Y}_a,y_{a,0})_{a\in[K]}$ \\
\textbf{Procedure}:
\begin{algorithmic}[1] 
\STATE Initialize $t \gets 1$, $B_1 \gets B$, $n_{a,0} \gets 0$ for each $a \in \mathcal{A}$ \\
\WHILE{$B_t > 0$}
\FOR{each arm $a \in \mathcal{A}$}
\STATE Sample $\tilde{\theta}_a^{(t)} \sim \mathcal{P}_a(y_{a,n_{a,t-1}})$ and $\tilde{R}_{a,i}^{(t)} \sim \mathcal{R}_a(\tilde{\theta}_a^{(t)})$ for $i=1,\dots,\left\lfloor B_t/c_a \right\rfloor$
\FOR{$i=1,\dots,\left\lfloor B_t/c_a \right\rfloor$}
\STATE $\tilde{y}_{a,n} \gets \mathcal{U}_a(\tilde{y}_{a,n-1},\tilde{R}_{a,n})$
\ENDFOR
\ENDFOR
\FOR{each $n_{1:K} \in \mathcal{N}^I(B_t)$}
\STATE $\tilde{\Gamma}[n_{1:K}] \gets \mathbb{E}_{\tilde{y}(n_{1:K})}[\max_a \frac{\mu_a(\theta_a)}{c_a}]$
\ENDFOR
\FOR{each $n_{1:K} \in \mathcal{N}^I(B_t)$}
\STATE $b \gets \sum_{a=1}^K c_a n_a$
\FOR{each $a \in \mathcal{A}$}
\IF{$B - b \geq c_a$}
\STATE $\tilde{r}^z[n_{1:K},a] \gets \mathbb{E}_{\tilde{y}(n_{1:K})}[ \mu_a(\theta_a)] + (B-b -c_a) \times (\tilde{\Gamma}[n_{1:K}]-\tilde{\Gamma}[n_{1:K}+e_a])$
\ELSE
\STATE $\tilde{r}^z[n_{1:K},a] \gets 0$
\ENDIF
\ENDFOR
\ENDFOR
\STATE $\tilde{M}[(0, \ldots, 0)] \gets 0$
\FOR{each $n_{1:K} \in \mathcal{N}^I(B_t) \setminus \{0\} $ in order with increasing $\sum_{a=1}^K n_a$}
\STATE $\tilde{M}[n_{1:K}] \gets \max_{a:n_a > 0}\{\tilde{M}[n_{1:K}-e_a]+ \tilde{r}^z[n_{1:K}-e_a,a]\}$
\STATE $\tilde{A}[n_{1:K}] \gets \argmax_{a:n_a > 0}\{\tilde{M}[n_{1:K}-e_a]+ \tilde{r}^z[n_{1:K}-e_a,a]\}$
\ENDFOR
\STATE $m_{1:K} \gets \argmax_{n_{1:K} \in \mathcal{N}^I(B_t)}\{\tilde{M}[n_{1:K}]\}$, $n \gets \sum_{a=1}^K m_a$
\WHILE{$n > 0$}
\STATE $\tilde{a}_{n}^\star \gets \tilde{A}[m_{1:K}]$
\STATE $m_{1:K} \gets m_{1:K} - e_{\tilde{a}_n^\star}$, $n \gets n-1$
\ENDWHILE
\STATE $A_{t} \gets \tilde{a}_1^\star$
\IF{$B_{t} < c_{A_{t}}$}
\STATE break
\ELSE
\STATE Play $A_{t}$, receive $r_{t}$, pay $c_{A_t}$ ($B_{t+1} \gets B_t - c_{A_t}$)
\STATE Update $y_{a,n_{a,t-1}+1} \gets \mathcal{U}_{a}(y_{a,n_{a,t-1}},r_{t})$, and $n_{a,t} \gets \begin{cases}n_{a,t-1}+1&\text{ for }a=A_t\\  n_{a,t-1}&\text{ for }a \ne A_t\end{cases}$
\ENDIF
\STATE $t \gets t+1$.
\ENDWHILE
\end{algorithmic}
\end{algorithm}
    
The pseudo-code is given in Algorithm \ref{alg:irs.vemax}. $\tilde{\Gamma}[n_{1:K}]$ corresponds to $W^\text{BTS}(B-b(n_{1:K})-c_a,y_{t}(n_{1:K},\omega))$ (line 11). 
And IRS.V-Emax calculates $r^z(n_{1:t},a,\omega)$ (line 14) and $ M(n_{1:K},\omega)$ (line 16--20). 
Since $|\mathcal{N}^I(B)| = O(T_{\text{max}}^K)$, this calculation requires $O(KT_{\text{max}}^K)$ computations.
We can obtain $W^\text{IRS.V-EMax}$ through the expected value of the above equation, and by tracking $M(n_{1:K},\omega)$'s backward, we can also obtain the optimal action sequence (line 21--31).

\subsection{IRS.INDEX Policy}
    
IRS.INDEX policy approximates the finite-horizon Gittins index \cite{kaufmann2012} together with the penalty function developed for IRS.V-EMax. 
    
First, we consider a single-armed bandit problem with an outside option.
We have a single stochastic arm $a$ incurring a cost $c_a$ and an outside option that also incurs the same cost $c_a$ but yields a deterministic reward $\lambda$.
According to \cite{min2019thompson}, the inner problem solved by IRS.V-EMax in this problem can be written as
\begin{equation*}
\begin{split}
\max_{n \in \{0,1,\ldots,T\} } \left\{T \times \Gamma_0^\lambda(\omega_a) + (T-n) \times (\lambda - \min_{0\leq i \leq n}\Gamma_i^\lambda (\omega_a)) + \sum_{i=1}^n (\hat{\mu}_{a,i-1}(\omega_a)-\Gamma_{i-1}^\lambda (\omega_a)) \right\},
\end{split}
\end{equation*}
where $T \triangleq \lfloor B/c_a \rfloor$ is the maximum number of plays, $\omega_a$ encodes all random realizations of arm $a$, $\Gamma_{n}^\lambda(\omega_a) \triangleq E_{y_{a,n}} [\max(\mu_a(\theta_a),\lambda)]$.
     The optimal solution $n^\star$ represents how many times the stochastic arm should be played if the DM knows the entire future but is penalized. If $n^\star \geq 1$, we interpret that the stochastic arm $a$ is worth trying against the deterministic value $\lambda$.
     
The values of $\Gamma_n^\lambda$'s can be calculated via closed-from expressions:
\begin{remark}
Define
        $$ \gamma^\lambda(\alpha_a,\beta_a) 
        		\triangleq \mathbb{E}_{\theta_a \sim \text{Beta}(\alpha_a, \beta_a)}\left[ \max\{ \theta_a, \lambda \} \right] 
		=  \lambda \times F_{\alpha_a,\beta_a}^{beta}(\lambda)+\frac{\alpha_a}{\alpha_a+\beta_a}\times (1 - F_{\alpha+1,\beta}^{beta}(\lambda)). $$
In the case of Beta-Bernoulli MAB, $\Gamma_n^\lambda(\omega_a) = \gamma^\lambda( \alpha_{a,0} + \sum_{i=1}^n R_{a,i},  \beta_{a,0} + n - \sum_{i=1}^n R_{a,i} )$.
And, in the case of Beta-Binomial MAB, $\Gamma_n^\lambda(\omega_a) = m_a \times \gamma^{\lambda/m_a}( \alpha_{a,0} + \sum_{i=1}^n R_{a,i},  \beta_{a,0} + n - \sum_{i=1}^n R_{a,i} )$.
\end{remark}

Motivated by the inner problem above, IRS.INDEX finds the threshold value $\lambda_a^\star$ beyond which the stochastic arm $a$ is not worth trying.
To explicitly obtain $\lambda_a^\star$, define $\varphi_a(\lambda,\omega_a)$, which represents the relative benefit of playing an arm $a$ versus when arm $a$ is not played at all:
\begin{equation*}
\small
\begin{split}
\varphi_a(\lambda,\omega_a) &\triangleq     \max_{n \in \{0,\ldots, T\}} \left\{T \times \Gamma_0^\lambda(\omega_a) + (T-n) \times (\lambda - \min_{0\leq i \leq n}\Gamma_i^\lambda(\omega_a)) + \sum_{i=1}^n (\hat{\mu}_{a,n_{i-1}}-\Gamma_{i-1}^\lambda(\omega_a)) \right\} - T \times \lambda.
\end{split}
\end{equation*}
Then, we can define $\lambda_a^\star$ as
$$\lambda_a^\star(\omega_a) \triangleq \sup\{\lambda \in \mathbb{R}; \varphi_a(\lambda,\omega_a) \geq 0\},$$
which is interpreted as the value of arm $a$ (given the outcome $\omega_a$), and is analogous to the Gittins index.
Given $\omega_a$, the value of $\lambda_a^\star(\omega_a)$ can be determined by the bisection search.
    
IRS.INDEX policy, just like the other IRS policies, utilizes the sampled outcome $\tilde{\omega}_a$ and computes $\tilde{\lambda}_a^\star(\tilde{\omega}_a)$, and then selects the arm with the highest value-to-cost ratio, i.e.,
$$ A_t \gets \argmax_{a \in \mathcal{A}} \frac{\lambda_a^*(\tilde{\omega}_a)}{c_a}. $$
The detailed procedure is implemented in Algorithm \ref{alg:irs.index} below.
\begin{algorithm}[h!]
\caption{IRS.INDEX}
\label{alg:irs.index}
\textbf{Input}: $K,B, (c_a,\mathcal{R}_a,\Theta_a,\mathcal{P}_a,\mathcal{Y}_a,y_{a,0})_{a\in[K]}$ \\
\textbf{Procedure}:
\begin{algorithmic}[1] 
\STATE Initialize $t \gets 1$, $B_1 \gets B$, $n_{a,0} \gets 0$ for each $a \in \mathcal{A}$ \\
\WHILE{$B_t > 0$}
\FOR{each arm $a \in \mathcal{A}$}
\STATE Sample $\tilde{\theta}_a^{(t)} \sim \mathcal{P}_a(y_{a,n_{a,t-1}})$ and $\tilde{R}_{a,i}^{(t)} \sim \mathcal{R}_a(\tilde{\theta}_a^{(t)})$ for $i=1,\dots,\left\lfloor B_t/c_a \right\rfloor$
\FOR{$i=1,\dots,\left\lfloor B_t/c_a \right\rfloor$}
\STATE $\tilde{y}_{a,n} \gets \mathcal{U}_a(\tilde{y}_{a,n-1},\tilde{R}_{a,n})$
\ENDFOR
\STATE $\lambda_a^\star \gets \sup\{\lambda \in \mathbb{R}; \varphi_a(\lambda,\tilde{\theta}_a,\tilde{R}_{a,1:n},\tilde{y}_{a,1:n}) \geq 0\}$ via bisection search
\ENDFOR
\STATE $A_{t} \gets \argmax_{a \in \mathcal{A}} \{ \lambda_a^\star / c_a \}$
\STATE Play $A_{t}$ and update variables (Algorithm \ref{alg:irs.vemax} lines 33--39)
\ENDWHILE
\end{algorithmic}
\end{algorithm}

\section{Extension to Random Costs}
\subsection{Simple extension}
The modified versions of IRS.FH and IRS.V-Zero  policy can be implemented in Algorithm \ref{alg:s-ext-irs.fh} and \ref{alg:s-ext-irs.vzero}.

\begin{algorithm}[htp!]
\caption{Simple extension of IRS.FH (IRS.FH.S-EXT)}
\label{alg:s-ext-irs.fh}
\textbf{Input}: $K,B, (\mathcal{R}_a,\Theta_a^{r},\mathcal{P}_a^{r},\mathcal{Y}_a^{r},y_{a,0}^{r},\mathcal{C}_a,\Theta_a^{c},\mathcal{P}_a^{c},\mathcal{Y}_a^{c},y_{a,0}^{c})_{a\in[K]}$ \\
\textbf{Procedure}:
\begin{algorithmic}[1] 
\STATE Initialize $t \gets 1$, $B_1 \gets B$, $n_{a,0} \gets 0$ for each $a \in \mathcal{A}$ \\
\WHILE{$B_t > 0$}
\FOR{each arm $a \in \mathcal{A}$}
\STATE Sample $\tilde{\theta}_a^{c,(t)} \sim \mathcal{P}_a^{c}(y_{a,n_{a,t-1}}^{c})$
\STATE Sample $\tilde{\theta}_a^{r,(t)} \sim \mathcal{P}_a^{r}(y_{a,n_{a,t-1}}^{r})$ and $\tilde{R}_{a,i}^{(t)} \sim \mathcal{R}_a(\tilde{\theta}_a^{r,(t)})$ for $i=1,\dots, \left\lfloor B_t/\mu(\tilde{\theta}_a^{c,(t)}) \right\rfloor $
\STATE $\tilde{\hat{\mu}}_{a,\left\lfloor B_t/\mu(\tilde{\theta}_a^{c,(t)}) \right\rfloor}^{r,(t)} \gets \hat{\mu}_{a, \left\lfloor B_t/\mu(\tilde{\theta}_a^{c,(t)}) \right\rfloor }( \tilde{R}_{a,1:\left\lfloor B_t/\mu(\tilde{\theta}_a^{c,(t)}) \right\rfloor}^{(t)} ; y_{a,n_{a,t-1}}^{r})$
\ENDFOR
\STATE Play $A_{t} \gets \argmax_{a \in \mathcal{A}} \{ \tilde{\hat{\mu}}_{a,\left\lfloor B_t/\mu(\tilde{\theta}_a^{c,(t)}) \right\rfloor}^{r,(t)} / \mu(\tilde{\theta}_a^{c,(t)})\}$, and observe $c_{t}$
\IF{$B_{t} < c_{t}$}
\STATE break
\ELSE
\STATE  Receive $r_{t}$, pay $c_{t}$ ($B_{t+1} \gets B_t - c_{t}$)
\STATE $y_{a,n_{a,t-1}+1}^r \gets \mathcal{U}_{a}(y_{a,n_{a,t-1}}^r,r_{t})$, $y_{a,n_{a,t-1}+1}^{r} \gets \mathcal{U}_{a}(y_{a,n_{a,t-1}}^c,c_{t})$, $n_{a,t} \gets n_{a,t-1}+1$ for $a=A_t$, \\ \ \ \ \ \ \ and $n_{a,t} \gets n_{a,t-1}$ for $a \ne A_t$
\ENDIF
\STATE $t \gets t+1$.
\ENDWHILE
\end{algorithmic}
\end{algorithm}

\begin{algorithm}[htp!]
\caption{Simple extension of IRS.V-Zero (IRS.V-Zero.S-EXT)}
\label{alg:s-ext-irs.vzero}
\textbf{Input}: $K,B, (\mathcal{R}_a,\Theta_a^{r},\mathcal{P}_a^{r},\mathcal{Y}_a^{r},y_{a,0}^{r},\mathcal{C}_a,\Theta_a^{c},\mathcal{P}_a^{c},\mathcal{Y}_a^{c},y_{a,0}^{c})_{a\in[K]}$ \\
\textbf{Procedure}:
\begin{algorithmic}[1] 
\STATE Initialize $t \gets 1$, $B_1 \gets B$, $n_{a,0} \gets 0$ for each $a \in \mathcal{A}$ \\
\WHILE{$B_t > 0$}
\FOR{each arm $a \in \mathcal{A}$}
\STATE Sample $\tilde{\theta}_a^{c,(t)} \sim \mathcal{P}_a^{c}(y_{a,n_{a,t-1}}^{c})$
\STATE Sample $\tilde{\theta}_a^{r,(t)} \sim \mathcal{P}_a^{r}(y_{a,n_{a,t-1}}^{r})$, and $\tilde{R}_{a,i}^{(t)} \sim \mathcal{R}_a(\tilde{\theta}_a^{r,(t)})$ for $i=1,\dots, \left\lfloor B_t/\mu(\tilde{\theta}_a^{c,(t)}) \right\rfloor $
\FOR{$i=1,\dots,\left\lfloor B_t/\mu(\tilde{\theta}_a^{c,(t)}) \right\rfloor$}
\STATE $\tilde{\hat{\mu}}_{a,i}^{r,(t)} \gets \hat{\mu}_{a,i}( \tilde{R}_{a,1:i}^{(t)} ; y_{a,n_{a,t-1}}^r)$
\ENDFOR
\ENDFOR
\STATE Solve $\tilde{n}_{1:K}^\star \gets \argmax_{\tilde{n}_{1:K} \in \mathcal{N}(B_t)} \sum_{a=1}^K \sum_{i=1}^{\tilde{n}_a} \tilde{\hat{\mu}}_{a,i-1}^{r,(t)}$ \\
\ \ \ where $\mathcal{N}(B_t) \triangleq \{(\tilde{n}_1,\dots,\tilde{n}_K);\sum_{a=1}^K \mu(\tilde{\theta}_a^{c,(t)}) \cdot \tilde{n}_a \leq B_t\}$
\STATE Play $A_{t}$ and update variables (Algorithm\ref{alg:s-ext-irs.fh} lines 9-15)
\ENDWHILE
\end{algorithmic}
\end{algorithm}
\newpage
\subsection{Extension with additional penalties}\label{subsec:p-ext}
The extension of IRS.V-zero explained in \S 3.5 is implemented in Algorithm \ref{alg:p-ext-irs.vzero}.
\begin{algorithm}[ht!]
\caption{Extension of IRS.V-Zero with additional penalties (IRS.V-Zero.P-EXT) }
\label{alg:p-ext-irs.vzero}
\textbf{Input}: $K,B, (\mathcal{R}_a,\Theta_a^{r},\mathcal{P}_a^{r},\mathcal{Y}_a^{r},y_{a,0}^{r},\mathcal{C}_a,\Theta_a^{c},\mathcal{P}_a^{c},\mathcal{Y}_a^{c},y_{a,0}^{c})_{a\in[K]}$\\
\textbf{Procedure}:
\begin{algorithmic}[1] 
\STATE Initialize $t \gets 1$, $B_1 \gets B$, $n_{a,0} \gets 0$ for each $a \in \mathcal{A}$ \\
\WHILE{$B_t > 0$}
\FOR{each arm $a \in \mathcal{A}$}
\STATE Sample $\tilde{\theta}_a^{c,(t)} \sim \mathcal{P}_a^{c}(y_{a,n_{a,t-1}}^{c})$ and $\tilde{C}_{a,i}^{(t)} \sim \mathcal{C}_a(\tilde{\theta}_a^{c,(t)})$ for $i=1,\dots, \text{ until } \sum_{i} \tilde{C}_{a,i}^{(t)} \leq B_t$
\STATE Sample $\tilde{\theta}_a^{r,(t)} \sim \mathcal{P}_a(y_{a,n_{a,t-1}}^{r})$ and $\tilde{R}_{a,i}^{(t)} \sim \mathcal{R}_a(\tilde{\theta}_a^{r,(t)})$ for $i=1,\dots,\text{ until } \sum_{i} \tilde{C}_{a,i}^{(t)} \leq B_t$
\WHILE{$\sum_i \tilde{C}_{a,i}^{(t)} \leq B_t$}
\STATE $\tilde{\hat{\mu}}_{a,i}^{r,(t)} \gets \hat{\mu}_{a,i}( \tilde{R}_{a,1:i}^{(t)} ; y_{a,n_{a,t-1}}^{r})$
\STATE $\tilde{\hat{\mu}}_{a,i}^{c,(t)} \gets \hat{\mu}_{a,i}( \tilde{C}_{a,1:i}^{(t)} ; y_{a,n_{a,t-1}}^{c})$
\ENDWHILE
\ENDFOR
\STATE $\tilde{\lambda} \gets \max_a \mu(\tilde{\theta}_a^{r,(t)}/\mu(\tilde{\theta}_a^{c,(t)})$ 
\STATE Solve $\tilde{n}_{1:K}^\star \gets \argmax_{\tilde{n}_{1:K} \in \mathcal{N}(B_t)} \sum_{a=1}^K \sum_{i=1}^{\tilde{n}_a} \Big\{\tilde{\hat{\mu}}_{a,i-1}^{r,(t)} + \tilde{\lambda}(\tilde{C}_{a,i}^{(t)} - \tilde{\hat{\mu}}_{a,i-1}^{r,(t)})\Big\}$ \\
	\ \ \ where $\mathcal{N}(B_t) \triangleq \{(\tilde{n}_1,\dots,\tilde{n}_K);\sum_{a=1}^K \sum_{i=1}^{\tilde{n}_a} \tilde{C}_{a,i}^{(t)} \leq B_t\}$
\STATE $A_{t} \gets \argmax_{a \in \mathcal{A}} \tilde{n}_a^\star$
\STATE Play $A_{t}$ and update variables (Algorithm\ref{alg:s-ext-irs.fh} lines 9-15)
\ENDWHILE
\end{algorithmic}
\end{algorithm}

We also extended the IRS.V-EMax and IRS.INDEX Policy with random cost settings. We describe how to extend the two algorithms with additional penalties.
\paragraph{IRS.V-EMax.P-EXT}
Recall the IRS.V-EMax. In random cost setting, since $c_t(a_{1:t},\omega) \triangleq C_{a_t,n_t(a_{1:t},a_t)}$, and $b_t(a_{1:t},\omega) \triangleq \sum_s^t c_s(a_{1:s},\omega)$
\begin{equation*}
\begin{split}
            \mathbb{E}\Big[W^\text{BTS}(B-b_{t}(a_{1:t},\omega),y_{t}(a_{1:t},\omega;y))|H_{t-1}\Big] =&  \mathbb{E}_y\Big[(B-b_{t}(a_{1:t},\omega)) \times \max_a \frac{\mu(\theta_a^r)}{\mu(\theta_a^c)}|H_{t-1}\Big] \\
            =&  \mathbb{E}_y\Big[(B-b_{t-1}(a_{1:t-1},\omega) - c_t(a_{1:t},\omega)) \times \max_a \frac{\mu(\theta_a^r)}{\mu(\theta_a^c)}|H_{t-1}\Big]
\end{split}
\end{equation*}
,where $y \triangleq (y^r,y^c)$, $\omega \triangleq (\theta^r,(R_{a,n})_{a\in \mathcal{A},n\in \mathbb{N}},\theta^c,(C_{a,n})_{a\in \mathcal{A},n\in \mathbb{N}}) \sim \mathcal{I}(y)$, and 

$H_{t-1}(a_{1:t-1},\omega) \triangleq (a_1,r_1(a_1,\omega),c_{t-1}(a_1,\omega),\dots, a_{t-1},r_{t-1}(a_{1:t-1},\omega),c_{t-1}(a_{1:t-1},\omega))$.
\newline
We approximate $\mathbb{E}\Big[W^\text{BTS}(B-b_{t}(a_{1:t},\omega),y_{t}(a_{1:t},\omega;y))|H_{t-1}\Big]$ as $$W^\text{BTS}(B-b_{t-1}(a_{1:t-1},\omega) - \hat{\mu}_{a_t,n_{t-1}(a_{1:t-1},a_t)}^c(\omega),y_{t-1}(a_{1:t-1},\omega;y))$$ for ease of computation, i.e.,
\begin{equation*}
\small
\begin{split}
            \mathbb{E}\Big[W^\text{BTS}(B-b_{t}(a_{1:t},\omega),y_{t}(a_{1:t},\omega;y))|H_{t-1}\Big]
            =&  \mathbb{E}_y\Big[(B-b_{t-1}(a_{1:t-1},\omega) - c_t(a_{1:t},\omega)) \times \max_a \frac{\mu(\theta_a^r)}{\mu(\theta_a^c)}|H_{t-1}\Big]\\
            \approx& (B-b_{t-1}(a_{1:t-1},\omega) - \hat{\mu}_{a_t,n_{t-1}(a_{1:t-1},a_t)}^c(\omega)) \times \mathbb{E}_{y_{t-1}} \Big[ \max_a \frac{\mu(\theta_a^r)}{\mu(\theta_a^c)}\Big]\\
            =& W^\text{BTS}(B-b_{t-1}(a_{1:t-1},\omega) - \hat{\mu}_{a_t,n_{t-1}(a_{1:t-1},a_t)}^c(\omega),y_{t-1}(a_{1:t-1},\omega;y))
\end{split}
\end{equation*}

Then, we can rewrite $z_t^\text{IRS.V-EMax}$ as
\begin{equation*}
\begin{split}
z_t^{\text{IRS.V-EMax}}\triangleq& r_t(a_{1:t},\omega) - \mathbb{E}_y\Big[r_t(a_{1:t},\omega)|H_{t-1}(a_{1:t-1},\omega)\Big]\\ 
&+ W^\text{BTS}(B-b_t(a_{1:t},\omega),y_{t}(a_{1:t},\omega;y))- \mathbb{E}_y\Big[W^\text{BTS}(B-b_t(a_{1:t},\omega),y_{t}(a_{1:t},\omega;y))|H_{t-1}(a_{1:t-1},\omega)\Big]\\
=&z_t^{\text{IRS.V-Zero}} + W^\text{BTS}(B-b_t(a_{1:t},\omega),y_{t}(a_{1:t},\omega;y)) \\
&-W^\text{BTS}(B-b_{t-1}(a_{1:t-1},\omega)- \hat{\mu}_{a_t,n_{t-1}(a_{1:t-1},a_t)}^c(\omega),y_{t-1}(a_{1:t-1},\omega;y)).
\end{split}
\end{equation*}

Let $y(n_{1:K},\omega)$ be the belief as a function of pull counts and $b(n_{1:K})\triangleq \sum_{a=1}^K \sum_{i=1}^{n_a} C_{a,i} $ be the consumed budget as a function of pull counts.
Then we can write the payoff from arm $a$ given the past number of plays $n_{1:K}$ in the inner problem solved by IRS.V-EMax as
\begin{equation*}
r^z(n_{1:K},a,\omega) \triangleq 
\left\{ \begin{array}{ll} 
            	
\begin{split}
&\hat{\mu}_{a,n_a}^r(\omega)-  W^\text{BTS}(B-b(n_{1:K}+e_a),y(n_{1:K}+e_a,\omega))\\&
\quad	+W^\text{BTS}(B-b(n_{1:K})-\hat{\mu}_{a,n_a}^c(\omega),y(n_{1:K},\omega)) 
\end{split} & \text{if } b(n_{1:K}+e_a) \leq B \\
0 & \text{if } b(n_{1:K}+e_a) > B.
\end{array} \right.
\end{equation*}
where $e_a \in \mathbb{N}_0^K$  is a basis vector such that the $a^{th}$ component is one and the others are zero.

And we denote the subproblem of the inner problem of IRS.V-EMax given $n_{1:K}$ as follows:
\begin{equation*}
\begin{split}
M(n_{1:K},\omega) \triangleq \max_{a_{1:t}\in \mathcal{A}^t}\Bigl\{\sum_{s=1}^{t} r_s(a_{1:s},\omega)- z_s^{\text{IRS.V-EMax}} ;\sum_{s=1}^t \textbf{1}\{a_s = a\} = n_a, \forall a\Bigl\},
\end{split}
\end{equation*}
where $t = \sum_{a=1}^K n_a$.
Then maximal value of $M(n_{1:K},\omega)$ must satisfy the following Bellman equation:
\begin{equation*}
\begin{split}
M(n_{1:K},\omega) = \max_{a \in \mathcal{A}} \Bigl\{ M(n_{1:K} - e_a,\omega) + r^z(n_{1:K}-e_a, a, \omega) \Bigl\}.
\end{split}
\end{equation*}
Therefore, the maximal value of the inner problem can be obtained by finding $n_{1:K}^\star$ that maximizes $M(n_{1:K},\omega)$, i.e.,
$$\max_{n_{1:K} \in \mathcal{N}^I(B)} \{M(n_{1:K},\omega)\}.$$

\begin{algorithm}[ht!]
\caption{IRS.V-EMax.P-EXT}
\label{alg:irs.vemax.p-ext}
\textbf{Input}: $K,B, (\mathcal{R}_a,\Theta_a^{r},\mathcal{P}_a^{r},\mathcal{Y}_a^{r},y_{a,0}^{r},\mathcal{C}_a,\Theta_a^{c},\mathcal{P}_a^{c},\mathcal{Y}_a^{c},y_{a,0}^{c})_{a\in[K]}$\\
\textbf{Procedure}:
\begin{algorithmic}[1] 
\STATE Initialize $t \gets 1$, $B_1 \gets B$, $n_{a,0} \gets 0$ for each $a \in \mathcal{A}$ \\
\WHILE{$B_t > 0$}
\FOR{each arm $a \in \mathcal{A}$}
\STATE Sample $\tilde{\theta}_a^{c,(t)} \sim \mathcal{P}_a^{c}(y_{a,n_{a,t-1}}^{c})$ and $\tilde{C}_{a,i}^{(t)} \sim \mathcal{C}_a(\tilde{\theta}_a^{c,(t)})$ for $i=1,\dots, \text{ until } \sum_{i} \tilde{C}_{a,i}^{(t)} \leq B_t$
\STATE Sample $\tilde{\theta}_a^{r,(t)} \sim \mathcal{P}_a(y_{a,n_{a,t-1}}^{r})$ and $\tilde{R}_{a,i}^{(t)} \sim \mathcal{R}_a(\tilde{\theta}_a^{r,(t)})$ for $i=1,\dots,\text{ until } \sum_{i} \tilde{C}_{a,i}^{(t)} \leq B_t$
\WHILE{$\sum_i \tilde{C}_{a,i}^{(t)} \leq B_t$}
\STATE $\tilde{y}_{a,n} \gets \mathcal{U}_a(\tilde{y}_{a,n-1},\tilde{R}_{a,n})$
\ENDWHILE
\ENDFOR
\FOR{each $n_{1:K} \in \mathcal{N}^I(B_t)$}
\STATE $\tilde{\Gamma}[n_{1:K}] \gets \mathbb{E}_{\tilde{y}(n_{1:K})}[\max_a \frac{\mu(\theta_a^r)}{\mu(\theta_a^c)}] \approx \mathbb{E}_{\tilde{y}(n_{1:K})}[\max_a \frac{\mu(\theta_a^r)}{\hat{\mu}_{a,n_a}^c}]$
\ENDFOR
\FOR{each $n_{1:K} \in \mathcal{N}^I(B_t)$}
\STATE $b \gets \sum_{a=1}^K\sum_{i=1}^{n_a} \tilde{C}_{a,i}^{(t)}$
\FOR{each $a \in \mathcal{A}$}
\IF{$B - b \geq c_a$}
\STATE $\tilde{r}^z[n_{1:K},a] \gets \mathbb{E}_{\tilde{y}(n_{1:K})}[ \mu(\theta_a^r)] +  ((B-b -\hat{\mu}_{a,n_a}^c(\omega)) \times \tilde{\Gamma}[n_{1:K}]- (B-b -\tilde{C}_{a,n_a+1}^{(t)})\times \tilde{\Gamma}[n_{1:K}+e_a])$
\ELSE
\STATE $\tilde{r}^z[n_{1:K},a] \gets 0$
\ENDIF
\ENDFOR
\ENDFOR
\STATE $\tilde{M}[(0, \ldots, 0)] \gets 0$
\FOR{each $n_{1:K} \in \mathcal{N}^I(B_t) \setminus \{0\} $ in order with increasing $\sum_{a=1}^K n_a$}
\STATE $\tilde{M}[n_{1:K}] \gets \max_{a:n_a > 0}\{\tilde{M}[n_{1:K}-e_a]+ \tilde{r}^z[n_{1:K}-e_a,a]\}$
\STATE $\tilde{A}[n_{1:K}] \gets \argmax_{a:n_a > 0}\{\tilde{M}[n_{1:K}-e_a]+ \tilde{r}^z[n_{1:K}-e_a,a]\}$
\ENDFOR
\STATE $m_{1:K} \gets \argmax_{n_{1:K} \in \mathcal{N}^I(B_t)}\{\tilde{M}[n_{1:K}]\}$, $n \gets \sum_{a=1}^K m_a$
\WHILE{$n > 0$}
\STATE $\tilde{a}_{n}^\star \gets \tilde{A}[m_{1:K}]$
\STATE $m_{1:K} \gets m_{1:K} - e_{\tilde{a}_n^\star}$, $n \gets n-1$
\ENDWHILE
\STATE $A_{t} \gets \tilde{a}_1^\star$
\STATE Play $A_{t}$ and update variables (Algorithm\ref{alg:s-ext-irs.fh} lines 9-15)
\ENDWHILE
\end{algorithmic}
\end{algorithm}

\paragraph{IRS.INDEX.P-EXT}
We consider a single-armed bandit problem with an outside option. We have a single stochastic arm $a$ incurring a stochastic cost $C_{a,t}$ and an outside option that also incurs the cost 1 but yields a deterministic reward $\lambda$. Then the inner problem solved by IRS.V-EMax.P-EXT in this problem can be written as
\begin{equation*}
    \begin{split}
        \text{maximize} &\sum_{t=1}^{\tau(a_{1:T_\text{max}})-1} \hat{\mu}_{a,n_{t-1}}^r(\omega_a) \cdot \textbf{1}\{a_t = 1\} + \lambda \cdot \textbf{1}\{a_t = 0\}\\
        &- \Big \{ (B-b_t(a_{1:t})) \cdot \Gamma_{n_t}^\lambda(\omega_a) - (B-b_{t-1}(a_{1:t-1})-\hat{\mu}_{a,n_{t-1}}^c(\omega_a)) \cdot \Gamma_{n_{t-1}}^\lambda(\omega_a) \Big \}\\
        \text{subject to } &n_t = \sum_{s=1}^t \textbf{1}\{a_t = 1\}, \quad \forall t = 1,\dots,T_\text{max},
    \end{split}
\end{equation*}
where $\hat{\mu}_{a,n_{t-1}}^r(\omega_a) \triangleq \mathbb{E}_{y_a^r}[\mu(\theta_a^r|R_{a,1},\dots,R_{a,n_{t-1}}]$, $\hat{\mu}_{a,n_{t-1}}^c(\omega_a) \triangleq \mathbb{E}_{y_a^c}[\mu(\theta_a^c|C_{a,1},\dots,C_{a,n_{t-1}}]$, and 
$$\Gamma_n^\lambda(\omega_a) \triangleq \mathbb{E}_{y_{a,n}}[\max(\frac{\mu(\theta_a^r)}{\mu(\theta_a^c)},\lambda)] \approx \mathbb{E}_{y_{a,n}}[\max(\frac{\mu(\theta_a^r)}{\hat{\mu}_{a,n}^c},\lambda)]$$

The values of $\Gamma_n^\lambda$'s can be calculated via closed-from expressions:
\begin{remark}
Define
\begin{equation*}
    \begin{split}
        \gamma^\lambda(\alpha_a^r,\beta_a^r,\hat{\mu}_{a,n}^c) 
        		&\triangleq \mathbb{E}_{\theta_a^r \sim \text{Beta}(\alpha_a^r, \beta_a^r)}\left[ \max\{ \frac{\mu(\theta_a^r)}{\hat{\mu}_{a,n}^c}, \lambda \} \right]\\
		&=  \lambda \times F_{\alpha_a^r,\beta_a^r,0,1/\hat{\mu}_{a,n}^c}^{beta}(\lambda)+\frac{\alpha_a^r}{(\alpha_a^r+\beta_a^r)\hat{\mu}_{a,n}^c}\times (1 - F_{\alpha_a^r+1,\beta_a^r,0,1/\hat{\mu}_{a,n}^c)}^{beta}(\lambda)).         
    \end{split}
\end{equation*}
In the case of Beta-Bernoulli MAB, $\Gamma_n^\lambda(\omega_a) = \gamma^\lambda( \alpha_{a,0} + \sum_{i=1}^n R_{a,i},  \beta_{a,0} + n - \sum_{i=1}^n R_{a,i},\hat{\mu}_{a,n}^c)$.
\end{remark}

Fix $m \triangleq n_{\tau-1}$, i.e., the total number of pulls on the stochastic arm. Assume that the stochastic arm is pulled first and then the deterministic arm is pulled, i.e., $a_t = 1$, for all $t=1,\dots,m$ and $a_t = 0$, for all $t=m+1,\dots,\tau-1$. Then note that if $a_t = 0$, then $\Big \{ (B-b_t(a_{1:t})) \cdot \Gamma_{n_t}^\lambda(\omega_a) - (B-b_{t-1}(a_{1:t-1})-\hat{\mu}_{a,n_{t-1}}^c(\omega_a)) \cdot \Gamma_{n_{t-1}}^\lambda(\omega_a) \Big \} = (1-\hat{\mu}_{a,n_t}^c) \cdot \Gamma_{n_t}^\lambda$ since $n_t = n_{t-1}$. The objective function can be represented as
$$\sum_{n=1}^m \hat{\mu}_{a,n-1}^r + (B-b_m)(\lambda - (1-\hat{\mu}_{a,m}^c) \cdot \Gamma_m^\lambda)-\sum_{n=1}^m \Big\{(B-b_n)\cdot\Gamma_n^\lambda - (B-b_{n-1}-\hat{\mu}_{a,n-1}^\lambda) \cdot \Gamma_{n-1}^\lambda\Big\}.$$
With $b_0 \triangleq 0$, we have
\begin{equation*}
    \begin{split}
        &\sum_{n=1}^m \Big\{(B-b_n)\cdot\Gamma_n^\lambda - (B-b_{n-1}-\hat{\mu}_{a,n-1}^\lambda) \cdot \Gamma_{n-1}^\lambda\Big\}\\
        =&\sum_{n=1}^m (B-b_n)\cdot\Gamma_n^\lambda -\sum_{n=1}^m (B-b_{n-1}-\hat{\mu}_{a,n-1}^\lambda) \cdot \Gamma_{n-1}^\lambda\\
        =&\sum_{n=1}^m (B-b_n)\cdot\Gamma_n^\lambda -\sum_{n=0}^{m-1} (B-b_{n}-\hat{\mu}_{a,n}^\lambda) \cdot \Gamma_{n}^\lambda\\
        =&\sum_{n=0}^m \Big\{(B-b_n)\cdot\Gamma_n^\lambda\Big\} -(B-b_0)\cdot \Gamma_0^\lambda -\sum_{n=0}^{m} \Big\{(B-b_{n}-\hat{\mu}_{a,n}^\lambda) \cdot \Gamma_{n}^\lambda\Big\} + (B-b_{m}-\hat{\mu}_{a,m}^\lambda) \cdot \Gamma_{m}^\lambda\\
        =& - B\cdot \Gamma_0^\lambda +(B-b_{m}-\hat{\mu}_{a,m}^\lambda) \cdot \Gamma_{m}^\lambda + \sum_{n=0}^m \hat{\mu}_{a,n}^\lambda \cdot \Gamma_{n}^\lambda\\
        =& - B\cdot \Gamma_0^\lambda +(B-b_{m}) \cdot \Gamma_{m}^\lambda + \sum_{n=1}^m \hat{\mu}_{a,n-1}^\lambda \cdot \Gamma_{n-1}^\lambda.
    \end{split}
\end{equation*}
So, the objective function can be reduced to 
\begin{equation*}
    \begin{split}
        &\sum_{n=1}^m \hat{\mu}_{a,n-1}^r + (B-b_m)(\lambda - (1-\hat{\mu}_{a,m}^c) \cdot \Gamma_m^\lambda)+ B\cdot \Gamma_0^\lambda -(B-b_{m}) \cdot \Gamma_{m}^\lambda - \sum_{n=1}^m \hat{\mu}_{a,n-1}^c \cdot \Gamma_{n-1}^\lambda\\
        &=B\cdot \Gamma_0^\lambda + (B-b_{m}) \cdot (\lambda-(2- \hat{\mu}_{a,m}^c)\cdot\Gamma_{m}^\lambda) + \sum_{n=1}^m  \Big\{\hat{\mu}_{a,n-1}^r -\hat{\mu}_{a,n-1}^c \cdot \Gamma_{n-1}^\lambda\Big\}
    \end{split}
\end{equation*}
Therefore, the inner problem can be reformulated as
$$\max_{n \in \{0,\ldots, \tau(a_{1:T_\text{max}})-1\}} \Big\{ B\cdot \Gamma_0^\lambda + (B-b_{n}) \cdot (\lambda-  (2- \hat{\mu}_{a,m}^c)\cdot \Gamma_{n}^\lambda) + \sum_{i=1}^n  \Big(\hat{\mu}_{a,i-1}^r -\hat{\mu}_{a,i-1}^c \cdot \Gamma_{i-1}^\lambda\Big)\Big\},$$
where the decision variable n is the total number of pulls on the stochastic arm.

Motivated by the inner problem above, IRS.INDEX finds the threshold value $\lambda_a^\star$ beyond which the stochastic arm $a$ is not worth trying.
To explicitly obtain $\lambda_a^\star$, define $\varphi_a(\lambda,\omega_a)$, which represents the relative benefit of playing an arm $a$ versus when arm $a$ is not played at all:
\begin{equation*}
\small
\begin{split}
\varphi_a(\lambda,\omega_a) &\triangleq     \max_{n \in \{0,\ldots, \tau(a_{1:T_\text{max}})-1\}} \left\{B\cdot \Gamma_0^\lambda + (B-b_{n}) \cdot (\lambda-(2- \hat{\mu}_{a,n}^c)\cdot \Gamma_{n}^\lambda) + \sum_{i=1}^n  \Big(\hat{\mu}_{a,i-1}^r -\hat{\mu}_{a,i-1}^c \cdot \Gamma_{i-1}^\lambda\Big)\right\} - B \times \lambda.
\end{split}
\end{equation*}
Then, we can define $\lambda_a^\star$ as
$$\lambda_a^\star(\omega_a) \triangleq \sup\{\lambda \in \mathbb{R}; \varphi_a(\lambda,\omega_a) \geq 0\},$$
which is interpreted as the value of arm $a$ (given the outcome $\omega_a$), and is analogous to the Gittins index.
Given $\omega_a$, the value of $\lambda_a^\star(\omega_a)$ can be determined by the bisection search.
IRS.INDEX policy, just like the other IRS policies, utilizes the sampled outcome $\tilde{\omega}_a$ and computes $\tilde{\lambda}_a^\star(\tilde{\omega}_a)$, and then selects the arm with the highest value, i.e.,
$$ A_t \gets \argmax_{a \in \mathcal{A}} \lambda_a^*(\tilde{\omega}_a). $$
The detailed procedure is implemented in Algorithm \ref{alg:irs.index.p-ext} below.

\begin{algorithm}[h!]
\caption{IRS.INDEX.P-EXT}
\label{alg:irs.index.p-ext}
\textbf{Input}: $K,B, (\mathcal{R}_a,\Theta_a^{r},\mathcal{P}_a^{r},\mathcal{Y}_a^{r},y_{a,0}^{r},\mathcal{C}_a,\Theta_a^{c},\mathcal{P}_a^{c},\mathcal{Y}_a^{c},y_{a,0}^{c})_{a\in[K]}$\\
\textbf{Procedure}:
\begin{algorithmic}[1] 
\STATE Initialize $t \gets 1$, $B_1 \gets B$, $n_{a,0} \gets 0$ for each $a \in \mathcal{A}$ \\
\WHILE{$B_t > 0$}
\FOR{each arm $a \in \mathcal{A}$}
\STATE Sample $\tilde{\theta}_a^{c,(t)} \sim \mathcal{P}_a^{c}(y_{a,n_{a,t-1}}^{c})$ and $\tilde{C}_{a,i}^{(t)} \sim \mathcal{C}_a(\tilde{\theta}_a^{c,(t)})$ for $i=1,\dots, \text{ until } \sum_{i} \tilde{C}_{a,i}^{(t)} \leq B_t$
\STATE Sample $\tilde{\theta}_a^{r,(t)} \sim \mathcal{P}_a(y_{a,n_{a,t-1}}^{r})$ and $\tilde{R}_{a,i}^{(t)} \sim \mathcal{R}_a(\tilde{\theta}_a^{r,(t)})$ for $i=1,\dots,\text{ until } \sum_{i} \tilde{C}_{a,i}^{(t)} \leq B_t$
\WHILE{$\sum_i \tilde{C}_{a,i}^{(t)} \leq B_t$}
\STATE $\tilde{y}_{a,n} \gets \mathcal{U}_a(\tilde{y}_{a,n-1},\tilde{R}_{a,n},\tilde{C}_{a,n})$
\ENDWHILE
\STATE $\lambda_a^\star \gets \sup\{\lambda \in \mathbb{R}; \varphi_a(\lambda,\tilde{\theta}_a^r,\tilde{R}_{a,1:n},\tilde{y}_{a,1:n}^r,\tilde{\theta}_a^c,\tilde{C}_{a,1:n},\tilde{y}_{a,1:n}^c) \geq 0\}$ via bisection search
\ENDFOR
\STATE $A_{t} \gets arg\max_{a \in \mathcal{A}} \{ \lambda_a^\star\}$
\STATE Play $A_{t}$ and update variables (Algorithm\ref{alg:s-ext-irs.fh} lines 9-15)
\ENDWHILE
\end{algorithmic}
\end{algorithm}

\section{Proof of Theorem 1}
The validity of bounds $W^\text{BTS}$, $W^\text{IRS.FH}$, $W^\text{IRS.V-Zero}$, and $W^\text{IRS.V-EMax}$ follows from the duality result (Theorem \ref{thm:duality}).
We provide the monotonicity among them.
    
\subsection{Proof of $W^\text{BTS}(B,y) \geq W^\text{IRS.FH}(B,y)$}
    
Since max() is a convex function, implying Jensen's inequality is as follows:
\begin{equation*}
\begin{split}
W^\text{BTS}(B,y) = B \times \mathbb{E}_y\Bigl[\max_a \frac{\mu_a(\theta_a)}{c_a}\Bigl]    \geq B \times \mathbb{E}_y\Bigl[\max_a \mathbb{E}\Bigl(\frac{\mu_a(\theta_a)}{c_a}\Bigl|\hat{\mu}_{a,\left \lfloor B/c_a \right \rfloor -1}\Bigl)\Bigl] =W^\text{IRS.FH}(B,y).
    \qquad \blacksquare
\end{split}
\end{equation*}
    
\subsection{Proof of $W^\text{IRS.FH}(B,y) \geq W^\text{IRS.V-Zero}(B,y)$}
    
We will utilize the following lemma.
\begin{lemma}[\cite{min2019thompson}]
On a probability space $(\Omega,\mathcal{F},\mathbb{P})$, let
$(c_0,\cdots,c_T)$ be $\mathcal{H}$-measurable real-valued random variables for some sub-$\sigma$-field $\mathcal{H} \subseteq \mathcal{F}$ (i.e., $c_i$'s are constants conditioned on $\mathcal{H}$). Then
\begin{equation*}
\begin{split}
\mathbb{E}\left[\max_{0\leq i \leq T }\{(i-n)^+ \times X + c_i\} \right] \geq \mathbb{E}\left[\max_{0\leq i \leq T }\{\mathbb{E}(X|\mathcal{H})\cdot 1\{i \geq n + 1\}+(i-n-1)^+ \times X + c_i\} \right].
\end{split}
\end{equation*}
\end{lemma}
    
Define $S_a(n_a) \triangleq \sum_{i=1}^{n_a} \hat{\mu}_{a,i-1}$.
What we want to show is
\begin{equation*}     \begin{split}
W^\text{IRS.FH.I} \equiv \mathbb{E}\Bigl[\max_{n_{1:K}\in \mathcal{N}^I(B)}\Bigl\{\sum_{a=1}^K n_a \times \hat{\mu}_{a,\left \lfloor B/c_a \right \rfloor -1} \Bigl\} \Bigl] \geq \mathbb{E}\Bigl[\max_{n_{1:K}\in \mathcal{N}^I(B)}\Bigl\{\sum_{a=1}^K S_a(n_a) \Bigl\} \Bigl] \equiv W^\text{IRS.V-Zero}.
\end{split}
\end{equation*}
Further define
\begin{equation*}
\small 
\begin{split}
    U_{k,n} \triangleq \mathbb{E}\Bigl[\max_{n_{1:K}\in \mathcal{N}^I(B)}\Bigl\{\Bigl(\sum_{a=1}^{k-1} S_a(n_a) \Bigl)+ (S_k (\min(n_k,n))+ (n_k-n)^+ \times \hat{\mu}_{a,\left \lfloor B/c_k \right \rfloor -1})+\Bigl(\sum_{a=k+1}^K n_a \times \hat{\mu}_{a,\left \lfloor B/c_a \right \rfloor -1} \Bigl) \Bigl\} \Bigl].
\end{split}
\end{equation*}
Observe that $W^\text{IRS.FH} = U_{1,0}, W^\text{IRS.V-Zero} = U_{K,\left \lfloor B/c_K \right \rfloor}$ and $U_{k+1,0} = U_{k,\left \lfloor B/c_k \right \rfloor} $.
Therefore, it suffices to show that
$$
    	U_{K,n}\geq U_{K,n+1}, \quad \forall k = 1,\cdots,K, \quad \forall n = 0,\cdots,\left \lfloor \frac{B}{c_k} \right \rfloor
$$
Fix $k$ and $n$, and define a sub-$\sigma$-field $\mathcal{H}$:
\[\mathcal{H} \triangleq \sigma(\{R_{a,s}\}_{a=k,1\leq s \leq n} \cup \{R_{a,s}\}_{a\ne k,1\leq s \leq \left \lfloor B/c_a \right \rfloor -1}). \]
For each $i = 0,\cdots,\left \lfloor B/c_k \right \rfloor$, define
\begin{equation*}
\begin{split}
D_{i,k} &\triangleq \max \Bigl \{ \Bigl( \sum_{a=1}^{k-1}S_a(n_a) \Bigl) + S_k(\min(i, n)) + \Bigl( \sum_{a=k+1}^K n_a \times \hat{\mu}_{a,\left \lfloor B/c_a \right \rfloor -1}\Bigl) : n_{1:K} \in \mathcal{N}^I(B), n_k = i\Bigl\}
\end{split}
\end{equation*}
Note that $D_{i,k}$'s are $\mathcal{H}$-measurable and 
\[U_{k,n} = \mathbb{E}\Bigl[\max_{0\leq i \leq \left \lfloor B/c_k \right \rfloor} \{(i-n)^+ \times \hat{\mu}_{a,\left \lfloor B/c_k \right \rfloor -1} +D_{i,k}\}\Bigl].\]
With $ X \triangleq \hat{\mu}_{a,\left \lfloor B/c_k \right \rfloor -1}$,
\begin{equation*}     \begin{split}
U_{k,n} =& \mathbb{E}\Bigl[\max_{0\leq i \leq \left \lfloor B/c_k \right \rfloor} \{(i-n)^+ \times \hat{\mu}_{a,\left \lfloor B/c_k \right \rfloor -1} +D_{i,k}\}\Bigl]\\
        &\overset{\underset{Lemma 1}{}}{\geq} \mathbb{E}\Bigl[\max_{0\leq i \leq \left \lfloor B/c_k \right \rfloor} \{\mathbb{E}(X|\mathcal{H})\cdot \textbf{1}\{i\geq n+1\}+(i-n-1)^+ \times X+D_{i,k}\}\Bigl]\\
        &\overset{\underset{(a)}{}}{=} \mathbb{E}\Bigl[\max_{0\leq i \leq \left \lfloor B/c_k \right \rfloor} \{\hat{\mu}_{k,n}\cdot \textbf{1}\{i\geq n+1\}+(i-n-1)^+\times \hat{\mu}_{a,n}+D_{i,k}\}\Bigl]\\
        &\overset{\underset{(b)}{}}{=} U_{k,n+1}.
\end{split}
\end{equation*}
Equation (a) holds since $\mathbb{E}(X|\mathcal{H}) = \mathbb{E}(\hat{\mu}_{a,\left \lfloor B/c_a \right \rfloor -1}|\mathcal{H}) =\mathbb{E}(\hat{\mu}_{a,\left \lfloor B/c_a \right \rfloor -1}|R_{k,1},\cdots, R_{k,n}) = \hat{\mu}_{a,n}$ and equation (b) holds since $S_k(\min(i, n))+ \hat{\mu}_{k,n} \cdot \textbf{1}\{i\geq n+1\} = \sum_{s=1}^n  \hat{\mu}_{k,s-1} \cdot \textbf{1}\{i\geq s\}+ \hat{\mu}_{k,n} \cdot \textbf{1}\{i\geq n+1\} = \sum_{s=1}^{n+1}  \hat{\mu}_{k,s-1} \cdot \textbf{1}\{i\geq s\} = S_k(\min(i, n+1))$.
Therefore, $W^\text{IRS.FH.I} \geq W^\text{IRS.V-Zero}$ and the claim follows from the fact that $W^\text{IRS.FH}  \geq W^\text{IRS.FH.I}$. $\blacksquare$
    
\subsection{Proof of $W^\text{BTS}(B,y) \geq W^\text{IRS.V-EMax}(B,y)$}
While not argued in the main paper, but we also have one more inequality $W^\text{BTS}(B,y) \geq W^\text{IRS.V-EMax}(B,y)$.
We here prove it.
\begin{definition}[Supersolution]
An approximate value function $\hat{V}:\mathbb{N}_0 \times \mathcal{Y} \mapsto \mathbb{R}$ is a \textbf{supersolution} to the Bellman equation if
    \[\hat{V}(B,y) \geq \max_{a \in \mathcal{A};B\geq c_a} \Bigl\{\mathbb{E}_{y_a} [R_{a,1}+\hat{V}(B-c_a,\mathcal{U}(y,R_{a,1},r))] \Bigl\}, \qquad \forall y \in \mathcal{Y}.\]
\end{definition}
    
\begin{remark}
If $\hat{V}(\cdot,\cdot,\cdot)$ is a supersolution, then for any given $\omega, B$ and y,
\begin{equation*}
\begin{split}
\hat{V}(B-b_t(a_{1:t})+1,y_{t-1}(a_{1:t-1},\omega;y)) \geq \mathbb{E}_y\Bigl[&r_t(a_{1:t-1}\oplus a,\omega;y)\\ &+ \hat{V}(B-b_t(a_{1:t-1}\oplus a),y_t(a_{1:t-1}\oplus a, \omega,y))|H_{t-1}(a_{1:t-1},\omega)\Bigl]
\end{split}
\end{equation*}
\end{remark}
\begin{lemma}\label{lem:supper_sol}
Consider a penalty function $\hat{z}_t$ generated by $\hat{V}(\cdot,\cdot,\cdot)$:
\begin{equation*}
\begin{split}
    \hat{z}_t(a_{1:t},\omega;B,y) \triangleq& r_t(a_{1:t},\omega) - \mathbb{E}_y[r_t(a_{1:t},\omega)|H_{t-1}(a_{1:t-1},\omega)]\\
    &+ \hat{V}(B-b_t(a_{1:t}),y_{t-1}(a_{1:t-1},\omega;y)) - \mathbb{E}_y\Bigl[ \hat{V}(B-b_t(a_{1:t}),y_t(a_{1:t}, \omega,y))|H_{t-1}(a_{1:t-1},\omega)\Bigl] 
\end{split}
\end{equation*}
If $\hat{V}(\cdot,\cdot,\cdot)$: is a supersolution, then the performance bound induced by penalty function $\hat{z}_t$ is tighter than $\hat{V}$: i.e.,
\[W^{\hat{z}}(B,y) \leq \hat{V}(B,y)\]
\end{lemma}
\textbf{Proof of $W^\text{BTS}(B,y) \geq W^\text{IRS.V-EMax}(B,y)$}
\newline
Recall that $z_t^{\text{IRS.V-EMax}}$ is a penalty function generated by $W^\text{BTS}$. We observe that $W^\text{BTS}(\cdot,\cdot,\cdot)$ is a supersolution: for any B and y,
\begin{equation*}
\begin{split}
    W^\text{BTS}(B,y)  =& \mathbb{E}_y \Bigl[B \times \max_{a \in \mathcal{A}} \frac{\mu_a(\theta_a)}{c_a} \Bigl]
    = \mathbb{E}_y \Bigl[\max_{a \in \mathcal{A}} \mu_a(\theta_a) + W^\text{BTS}(B-c_{a},y)\Bigl]\\
    \geq& \max_{a \in \mathcal{A}} \Bigl\{ \mathbb{E}_y [ \mu_a(\theta_a) + W^\text{BTS}(B-c_{a},y) ]\Bigl\}
    = \max_{a \in \mathcal{A}} \Bigl\{ \mathbb{E}_y [ R_{a,1}(\theta_a)] + W^\text{BTS}(B-c_{a},y)] \Bigl\}\\
    =& \max_{a \in \mathcal{A}} \Bigl\{ \mathbb{E}_y [ R_{a,1}(\theta_a)] + W^\text{BTS}(B-c_{a},\mathcal{U}(y,a,R_{a,1}))] \Bigl\}
\end{split}     \end{equation*}
By Lemma \ref{lem:supper_sol}, we deduce that $W^\text{BTS}(B,y) \geq W^\text{IRS.V-EMax}(B,y)$ which also holds in a stronger sense.  $\blacksquare$
   
\section{Proof of Theorem 2}
Without loss of generality, we assume that there is a \emph{null arm} that can be played even if there is no budget left: i.e., if $t \geq \tau(A_{1:T_{\text{max}}}^\pi)$ then $A_{t} = null$, $\mu_{null} = 0$, and $c_{null} = 1$.
We define the following to simplify the expressions:
$$N_{t-1}^\pi(a) \triangleq n_{t-1}(A_{1:t-1}^{\pi},a) , \qquad \hat{\mu}_{t}^\pi(a,n) \triangleq \hat{\mu}_{a,N_{t-1}^\pi(a)+n}. $$
Observe that for each $a \in \mathcal{A}$, the process $\{\hat{\mu}_{t}^\pi(a,n)\}_{n \geq 0}$ is martingale.
    
Further define
$$\triangle_{t}^\pi(a,n) \triangleq \sqrt{\frac{L}{\nu + N_{t-1}^\pi(a)} \times \frac{n}{\nu + N_{t-1}^\pi(a)+n}}, $$
which is measurable with respect to $\mathcal{F}_{t-1}$.
The prior/posterior of arm $a$ at time $t$ is described by the hyperparameters $\Bigl(\xi_a + \sum_{i=1}^{N_{t-1}^\pi(a)} R_{a,i} , \nu + N_{t-1}^\pi(a)\Bigl)$ that converges to $\mu_a$, and therefore Lemma 5 in \cite{min2019thompson} implies that $\hat{\mu}_{t}^\pi(a,\infty)$ is $\triangle_{t}^\pi(a,n)$-sub-Gaussian conditioned on $\mathcal{F}_{t-1}$.
    
We first start by observe the following.
    
\begin{remark}\label{remark: prob matching}
For each of penalty functions in section 2.6, the IRS policy $\pi$ is randomized in such a way that it takes an action $a$ with the probability that the action $a$ is indeed the best action $a_t^{z,*}$ at that moment, i.e.,
\[ \mathbb{P}[A_t^{\pi}=a|\mathcal{F}_{t-1}] = \mathbb{P}[a_t^{z,*}(A_{1:t-1}^{\pi})=a|\mathcal{F}_{t-1}] \quad \forall a , \forall t\]
The source of uncertainty in the LHS is the randomness of the policy (embedded in $\Tilde{\omega}$) and that in the RHS is the randomness of nature (embedded in $\omega$). Here we assume that the tie-breaking rule in argmax of $a_t^{z,*}(a_{1:t-1},\omega;B,y) \triangleq \argmax_{a \in \mathcal{A}}\{Q_t^{z,in}(a_{1:t-1},a,\omega;B,y)\}$ is identical to the one used when $\pi^z$ solves the inner problem.
\end{remark}
    
\subsection{BTS}\label{sec:subopt of BTS}
Let $\pi = \pi^\text{BTS}$, and $A^\star = \argmax_{a \in \mathcal{A}} \frac{\mu_a(\theta_a)}{c_a}$. Then we can decompose suboptimality gap of BTS and write it as:
    
\begin{equation*}
\begin{split}
            W^\text{BTS}(B,y) - V(\pi^\text{BTS},B,y)  
            &= \mathbb{E}_y \Bigl[B \times \max_{a \in \mathcal{A}} \frac{\mu_a}{c_a} - \sum_{t=1}^{\tau(A_{1:T_{\text{max}}}^{\pi})-1} \mu_{A_t^\pi} \Bigl]
            \\&= \mathbb{E}_y \Bigl[B \times \frac{\mu_{A^*}}{c_{A^*}} - \sum_{a=1}^K c_{A_t^\pi} \times \frac{\mu_{A_t^\pi}}{c_{A_t}^\pi} \Bigl]
            \\&= \mathbb{E}_y \Bigl[B \times \frac{\mu_{A^*}}{c_{A^*}} - \sum_{b=1}^{B} \frac{\mu_{A_{t^\pi(b)}^\pi}}{c_{A_{t^\pi(b)}^\pi}} \Bigl] 
            \\&= \mathbb{E}_y \Bigl[\sum_{b=1}^{B} (\frac{\mu_{A^*}}{c_{A^*}} - \frac{\mu_{A_{t^\pi(b)}^\pi}}{c_{A_{t^\pi(b)}^\pi}}) \Bigl],
\end{split}
\end{equation*}
where $t^\pi(b)$ denotes the (random) time index at which the $b^\text{th}$ unit of resource is being used i.e., $t^\pi(b) \triangleq min\{t:b_t(A_{1:t}^\pi \geq b\}$.
Define the sequence of confidence intervals:
$$L_{t}(a) \triangleq  \hat{\mu}_{t}^\pi(a,0) - \sqrt{2\log (\frac{B}{c_a})} \times \triangle_{t}^\pi(a,\infty), \quad U_{t}(a) \triangleq  \hat{\mu}_{t}^\pi(a,0) +\sqrt{2\log (\frac{B}{c_a})} \times \triangle_{t}^\pi(a,\infty).$$
where $\triangle_{t}^\pi(a,\infty) = \lim_{n \to \infty} \triangle_{t}^\pi(a,\infty) = \sqrt{\frac{L}{\nu+ N_{t-1}^\pi(a)}}$ so that $\mu_a$ is $\triangle_{t}^\pi(a,\infty)$-sub-Gaussian conditioned on $\mathcal{F}_{t-1}$.
\newline
By Lemma 4 in \cite{min2019thompson}, we have
$$\mathbb{E}\Bigl[(\mu_a - U_t(a))^+ | \mathcal{F}_{t-1} \Bigl] \leq \frac{\triangle_{t}^\pi(a,\infty)}{\sqrt{2\log (\frac{B}{c_a})}} e^{-\frac{2\log (\frac{B}{c_a})}{2}} \leq \frac{\sqrt{L/\nu}}{B/c_{a}}, $$
$$\mathbb{E}\Bigl[(L_t(a)- \mu_a)^+ | \mathcal{F}_{t-1} \Bigl] \leq \frac{\triangle_{t}^\pi(a,\infty)}{\sqrt{2\log (\frac{B}{c_a})}} e^{-\frac{2\log (\frac{B}{c_a})}{2}} \leq \frac{\sqrt{L/\nu}}{B/c_{a}}. $$
where we use the fact that $2\log (B/c_a)\geq 1 \text{ for any } B \geq 2c_{max}$.
Let $A_{t}^* \triangleq a_t^{z,*}(A_{1:t}^\pi,\omega)$ and $\mathbb{E}_t[\cdot]$ denote $\mathbb{E}_t[\cdot|\mathcal{F}_{t-1}]$ then by Remark \ref{remark: prob matching} we have
$$\mathbb{E}_t\Bigl[\frac{U_t(A_t^\pi)}{c_{A_t^\pi}}\Bigl] = \sum_{a \in \mathcal{A}} \frac{U_t(a)}{c_a} \cdot \mathbb{P}_t[A_t^\pi = a] = \sum_{a \in \mathcal{A}} \frac{L_t(a)}{c_a} \cdot \mathbb{P}_t[A_t^* = a] =\mathbb{E}_t\Bigl[\frac{U_t(A_t^*)}{c_{A_t^*}}\Bigl].$$
Therefore,
\begin{equation*}
\begin{split}
\mathbb{E}_t\Bigl[\frac{\mu_{A_{t}^*}}{c_{A_{t}^*}} - \frac{\mu_{A_{t}^\pi}}{c_{A_{t}^\pi}}\Bigl] 
            &= \mathbb{E}_t\Bigl[\frac{\mu_{A_{t}^*}}{c_{A_{t}^*}}  - \frac{\mu_{A_{t}^\pi}}{c_{A_{t}^\pi}} + \frac{U_t(A_{t}^\pi)}{c_{A_{t}^\pi}} -\frac{U_{t}(A_{t}^\pi)}{c_{A_{t}^\pi}}+\frac{L_{t}(A_{t}^\pi)}{c_{A_{t}^\pi}} -\frac{L_t(A_{t}^\pi)}{c_{A_{t}^\pi}} \Bigl]
            \\&= \mathbb{E}_t\Bigl[\frac{\mu_{A_{t}^*}}{c_{A_{t}^*}}  -\frac{U_t(A_{t}^*)}{c_{A_{t}^*}}  + \frac{U_t(A_{t}^\pi)}{c_{A_{t}^\pi}} -\frac{L_t(A_{t}^\pi)}{c_{A_{t}^\pi}} +\frac{L_t(A_{t}^\pi)}{c_{A_{t}^\pi}} - \frac{\mu_{A_{t}^\pi}}{c_{A_{t}^\pi}}\Bigl]
            \\&= \mathbb{E}_t\Bigl[\frac{(\mu_{A_{t}^*}-U_t(A_{t}^*))}{c_{A_{t}^*}}   +\frac{(L_t(A_{t}^\pi)-\mu_{A_{t}^\pi})}{c_{A_{t}^\pi}}\ + \frac{(U_t(A_{t}^\pi)-L_t(A_{t}^\pi))}{c_{A_{t}^\pi}}\Bigl]\\
             &\leq \mathbb{E}_t\Bigl[\frac{(\mu_{A_{t}^*}-U_t(A_{t}^*))^+}{c_{A_{t}^*}}\Bigl] +\mathbb{E}_t\Bigl[\frac{(L_t(A_{t}^\pi)-\mu_{A_{t}^\pi})^+}{c_{A_{t}^\pi}}\Bigl] + \mathbb{E}_t\Bigl[\frac{(U_t(A_{t}^\pi)-L_t(A_{t}^\pi))}{c_{A_{t}^\pi}} \Bigl].
\end{split}
\end{equation*}
We further observe that, by Remark \ref{remark: prob matching},
\begin{equation*}
\begin{split}
            \mathbb{E}_t\Bigl[\frac{(\mu_{A_{t}^*}-U_t(A_{t}^*))^+}{c_{A_{t}^*}}\Bigl]  =&\sum_{a \in \mathcal{A}} \mathbb{E}_t\Bigl[\frac{(\mu_{A_{t}^*}-U_t(A_{t}^*))^+}{c_{A_{t}^*}}\Bigl] \mathbb{P}_t[A_t^\pi = a]= \sum_{a \in \mathcal{A}} \mathbb{E}_t\Bigl[\frac{(\mu_{A_{t}^*}-U_t(A_{t}^*))^+}{c_{A_{t}^*}}\Bigl] \mathbb{P}_t[A_{t}^* = a] \\
            \leq& \frac{\sqrt{L/\nu}}{B/c_{A_t^*} \cdot c_{A_t^*}} \sum_{a \in \mathcal{A}} \mathbb{P}_t[A_{t}^* = a] \leq \frac{\sqrt{L/\nu}}{B}.
\end{split}
\end{equation*}
Similarly we have $\mathbb{E}_t\Bigl[\frac{(L_t(A_{t}^\pi) - \mu_{A_{t}^\pi})^+}{c_{A_{t}^\pi}}\Bigl] \leq \frac{\sqrt{L/\nu}}{B}$.
Combining these results, we deduce that
\begin{equation*}
\begin{split}
W^\text{BTS}(B,y) - V(\pi^{BTS},B,y) 
             &\leq  \mathbb{E}_y \Bigl[\sum_{b=1}^{B} (\frac{\mu_{A^*}}{c_{A^*}} - \frac{\mu_{A_{t^\pi(b)}^\pi}}{c_{A_{t^\pi(b)}^\pi}}) \Bigl] 
             =  \mathbb{E}_y \Bigl[\sum_{b=1}^{B} \mathbb{E}_{t^\pi(b)}\Bigl[\frac{\mu_{A^*}}{c_{A^*}} - \frac{\mu_{A_{t^\pi(b)}^\pi}}{c_{A_{t^\pi(b)}^\pi}}\Bigl] \Bigl]
            \\&\leq  \mathbb{E}_y \Bigl[\sum_{b=1}^{B} \frac{2\sqrt{L/\nu}}{B} + \mathbb{E}_{t^\pi(b)}\Bigl[\frac{(U_{t^\pi(b)}(A_{t^\pi(b)}^\pi)-L_{t^\pi(b)}(A_{t^\pi(b)}^\pi))}{c_{A_{t^\pi(b)}^\pi}} \Bigl] \Bigl]
            \\&=  2\sqrt{L/\nu} +\mathbb{E}_y \Bigl[\sum_{b=1}^{B} \mathbb{E}_{t^\pi(b)}\Bigl[\frac{(U_{t^\pi(b)}(A_{t(b)}^\pi)-L_{t^\pi(b)}(A_{t^\pi(b)}^\pi))}{c_{A_{t^\pi(b)}^\pi}}\Bigl]\Bigl] 
            \\&=  2\sqrt{L/\nu} +  \mathbb{E}_y \Bigl[ \sum_{b=1}^{B} 2\sqrt{2\log(\frac{B}{c_{A_{t^\pi(b)}^\pi}})} \times  \frac{\triangle_{t^\pi(b)}^\pi(A_{t^\pi(b)}^\pi,\infty)}{c_{A_{t^\pi(b)}^\pi}}\Bigl]
            \\&\leq  2\sqrt{L/\nu}  + 2\sqrt{2\log T_{\text{max}}} \times \mathbb{E}_y \Bigl[\sum_{b=1}^{B} \frac{\triangle_{t^\pi(b)}^\pi(A_{t^\pi(b)}^\pi,\infty)}{c_{A_{t^\pi(b)}^\pi}}\Bigl].
\end{split}
\end{equation*}
We can establish an upper bound on the term $\mathbb{E}_y \Bigl[\sum_{b=1}^{B} \frac{\triangle_{t^\pi(b)}^\pi(A_{t^\pi(b)}^\pi,\infty)}{c_{A_{t^\pi(b)}^\pi}}\Bigl]$ as follows:
\begin{equation*}
\begin{split}
            \sum_{b=1}^{B}\frac{\triangle_{t^\pi(b)}^\pi(A_{t^\pi(b)}^\pi,\infty)}{c_{A_{t^\pi(b)}^\pi}} =& \sum_{b=1}^{B} \frac{1}{c_{A_{t^\pi(b)}^\pi}}\sqrt{\frac{L}{\nu+N_{t(b)-1}^\pi(A_{t(b)}^\pi)}} \leq \sum_{a \in \mathcal{A}} \sum_{n=0}^{N_{t^\pi(B)-1}^\pi(a)-1} \sqrt{\frac{L}{\nu+ n}}\\
& = \sum_{a \in \mathcal{A}} \Bigl(\sqrt{\frac{L}{\nu}}+ \sum_{n=1}^{N_{t^\pi(B)-1}^\pi(a)-1} \sqrt{\frac{L}{\nu+ n}}\Bigl)
            \leq \sum_{a \in \mathcal{A}}  \Bigl(\sqrt{\frac{L}{\nu}}+ \sum_{n=1}^{N_{t^\pi(B)-1}^\pi(a)-1} \sqrt{\frac{L}{n}}\Bigl)\\
            &\leq \sum_{a \in \mathcal{A}}  \Bigl(\sqrt{\frac{L}{\nu}}+ \int_{n=1}^{N_{t^\pi(B)-1}^\pi(a)} \sqrt{\frac{L}{x}} dx\Bigl)
            \leq K\frac{\sqrt{L}}{\sqrt{\nu}} + 2\sqrt{L}\sum_{a \in \mathcal{A}} \sqrt{N_{t^\pi(B)}^\pi (a)}.
\end{split}
\end{equation*}
By utilizing Cauchy-Schwartz inequality, we deduce that
\begin{equation*}
\begin{split}
            \sum_{a \in \mathcal{A}} \sqrt{N_{t^\pi(B)}^\pi (a)} \leq \sqrt{K\sum_{a\in \mathcal{A}} N_{t^\pi(B)}(a)} \leq \sqrt{KT_{\text{max}}}.
\end{split}
\end{equation*}
Combining all these results,
\begin{equation*}
\begin{split}
            W^\text{BTS}(B,y) - V(\pi^{BTS},B,y) \leq&   2\sqrt{L/\nu} + 2\sqrt{2\log T_{\text{max}} } \times (K\frac{\sqrt{L}}{\sqrt{\nu}} + 2\sqrt{L} \sqrt{KT_{\text{max}}})\\
            \leq&  2\sqrt{L}\Bigl[\frac{1}{\sqrt{\nu}} + \sqrt{2\log T_{\text{max}}}\bigl(\frac{K}{\sqrt{\nu}} + 2 \sqrt{KT_{\text{max}}}\bigl)\Bigl].
\end{split}
\end{equation*}

\subsection{IRS.FH}\label{sec:subopt of FH}
Let $\pi = \pi^\text{IRS.FH}$, and $A^* = \argmax_{a \in \mathcal{A}} \frac{\hat{\mu}_{a,\left \lfloor B/c_a \right \rfloor -1}}{c_a}$. Then, like BTS, suboptimality gap of IRS.FH can be decomposed as follows.
\begin{equation*}
\begin{split}
W^\text{IRS.FH}(B,y) - V(\pi,B,y) 
= \mathbb{E}_y \left[\sum_{b=1}^{B} \frac{\hat{\mu}_{A^*,\left \lfloor B/c_{A^*} \right \rfloor -1}}{c_{A^*}} - \frac{\hat{\mu}_{A_{t^\pi(b)}^\pi,\left \lfloor B/c_{A_{t^\pi(b)}^\pi} \right \rfloor -1}}{c_{A_{t^\pi(b)}^\pi}} \right].
\end{split}
\end{equation*}
Define the sequence of confidence intervals:
$$L_{t}(a) \triangleq  \hat{\mu}_{t}^\pi(a,0) - \sqrt{2\log (B/c_a)} \times \triangle_{t}^\pi\left(a,\left\lfloor (B-b_{t-1}(A_{1:t-1}^\pi)) / c_a \right\rfloor \right) $$
$$ U_t(a) \triangleq  \hat{\mu}_{t}^\pi(a,0) + \sqrt{2\log (B/c_a)} \times \triangle_{t}^\pi\left(a,\left\lfloor (B-b_{t-1}(A_{1:t-1}^\pi))/c_a \right\rfloor \right) $$
Given that $\hat{\mu}_t^\pi(a,\left\lfloor\frac{B-b_t(A_{1:t}^\pi)}{c_a}\right\rfloor)$ is $\triangle_{t}^\pi(a,\left\lfloor\frac{B-b_t(A_{1:t}^\pi)}{c_a}\right\rfloor)$-sub-Gaussian conditioned on $\mathcal{F}_{t-1}$.
By Lemma 4 in \cite{min2019thompson}, we have
\begin{equation*}
\small
\begin{split}
\mathbb{E}\Bigl[(\hat{\mu}_t^\pi(a,\left\lfloor\frac{B-b_t(A_{1:t}^\pi)}{c_a}\right\rfloor)- U_t(a))^+ | \mathcal{F}_{t-1} \Bigl] 
\leq \frac{\triangle_{t}^\pi(a,\left\lfloor\frac{B-b_t(A_{1:t}^\pi)}{c_a}\right\rfloor)}{\sqrt{2\log (\frac{B}{c_a})}} e^{-\frac{2\log (\frac{B}{c_a})}{2}}
\leq \frac{\triangle_{t}^\pi(a,\infty)}{\sqrt{2\log (\frac{B}{c_a})}} e^{-\frac{2\log (\frac{B}{c_a})}{2}} \leq \frac{\sqrt{L/\nu}}{B/c_{a}}.
\end{split}
\end{equation*}
Similarly, we have
$$\mathbb{E}\Bigl[(L_t(a)- \hat{\mu}_t^\pi(a,\left\lfloor\frac{B-b_t(A_{1:t}^\pi)}{c_a}\right\rfloor)^+ | \mathcal{F}_{t-1} \Bigl] \leq \frac{\sqrt{L/\nu}}{B/c_{a}}, $$
and
\begin{equation*}
\begin{split}
&W^\text{IRS.FH}(B,y) - V(\pi,B,y)  \leq 2\sqrt{L/\nu}  + 2\sqrt{2\log T_{\text{max}}} \times \mathbb{E}_y \Bigl[ \sum_{b=1}^{B} \frac{\triangle_{t^\pi(b)}^\pi(A_{t^\pi(b)}^\pi,\left\lfloor\frac{B-b_{t^\pi(b)}(A_{1:t^\pi(b)}^\pi)}{c_a}\right\rfloor)}{c_{A_{t^\pi(b)}^\pi}}\Bigl].
\end{split}
\end{equation*}
On the other hand, since $N_{t^\pi(b)-1}(a) \leq \frac{b_{t^\pi(b)}(A_{1:t^\pi(b)}^\pi)}{c_\text{min}}$ in any case, we have
\begin{equation*}
\footnotesize
\begin{split}
            \frac{1}{\nu + N_{t^\pi(b)-1}^\pi(a)} \times \frac{\left\lfloor\frac{B-b_{t^\pi(b)}(A_{1:t^\pi(b)}^\pi)}{c_a}\right\rfloor}{\nu + N_{t^\pi(b)-1}^\pi(a)+\left\lfloor\frac{B-b_{t^\pi(b)}(A_{1:t^\pi(b)}^\pi)}{c_a}\right\rfloor} =& \frac{1}{\nu + N_{t^\pi(b)-1}^\pi(a)} \times \Bigl(1- \frac{\nu + N_{t^\pi(b)-1}^\pi(a)}{\nu + N_{t^\pi(b)-1}^\pi(a)+\left\lfloor\frac{B-b_{t^\pi(b)}(A_{1:t^\pi(b)}^\pi)}{c_a}\right\rfloor}\Bigl)\\
            =& \frac{1}{\nu + N_{t^\pi(b)-1}^\pi(a)} - 
            \frac{1}{\nu + N_{t^\pi(b)-1}^\pi(a)+\left\lfloor\frac{B-b_{t^\pi(b)}(A_{1:t^\pi(b)}^\pi)}{c_a}\right\rfloor}\\
            \leq& \frac{1}{\nu + N_{t^\pi(b)-1}^\pi(a)} - 
            \frac{1}{\nu + N_{t^\pi(b)-1}^\pi(a)+\frac{B-b_{t^\pi(b)}(A_{1:t^\pi(b)}^\pi)}{c_\text{min}}}\\
            \leq& \frac{1}{\nu + N_{t^\pi(b)-1}^\pi(a)} - \frac{1}{\nu +T_{\text{max}}}.
\end{split}
\end{equation*}
Consequently,
\begin{equation*}
\begin{split}
            \sum_{b=1}^{B} \frac{\triangle_{t^\pi(b)}^\pi(A_{t^\pi(b)}^\pi,\left\lfloor\frac{B-b_{t^\pi(b)}(A_{1:t^\pi(b)}^\pi)}{c_a}\right\rfloor)}{c_{A_{t^\pi(b)}^\pi}}
            =&\sqrt{L}\sum_{b=1}^{B} \frac{1}{c_{A_{t^\pi(b)}^\pi}} \sqrt{\frac{1}{\nu + N_{t^\pi(b)-1}^\pi(a)} - \frac{1}{\nu +T_{\text{max}}}}\\
            \leq& \sqrt{L}\sum_{a \in \mathcal{A}} \sum_{n=0}^{N_{t^\pi(B)-1}^\pi(a)-1} \sqrt{\frac{1}{\nu +n} - \frac{1}{\nu +T_{\text{max}}}}\\
            =& \sqrt{L}\sum_{a \in \mathcal{A}}\Bigl( \sqrt{\frac{1}{\nu}-\frac{1}{\nu+T_{\text{max}}}}+\sum_{n=1}^{N_{t^\pi(B)-1}^\pi(a)-1} \sqrt{\frac{1}{\nu +n} - \frac{1}{\nu +T_{\text{max}}}}\Bigl)\\
            \leq& \sqrt{L}\Bigl(\frac{K}{\sqrt{\nu}}+\sum_{a \in \mathcal{A}}\sum_{n=1}^{N_{t^\pi(B)-1}^\pi(a)-1} \sqrt{\frac{1}{n}-\frac{1}{T_{\text{max}}}}\Bigl)\\
            &\overset{\underset{(a)}{}}{\leq} \sqrt{L}\Bigl(\frac{K}{\sqrt{\nu}}+\sum_{a \in \mathcal{A}}\sum_{n=1}^{N_{t^\pi(B)-1}^\pi(a)-1} \Bigl(\frac{1}{\sqrt{n}}-\frac{\sqrt{n}}{2T_{\text{max}}}\Bigl)\Bigl)\\
            \leq& \sqrt{L}\Bigl(\frac{K}{\sqrt{\nu}}+\sum_{a \in \mathcal{A}}\int_{0}^{N_{t^\pi(B)-1}^\pi(a)} \Bigl(\frac{1}{\sqrt{x}}-\frac{\sqrt{x}}{{2T_{\text{max}}}}\Bigl)dx\Bigl)\\
            =& \sqrt{L}\Bigl(\frac{K}{\sqrt{\nu}}+ \sum_{a \in \mathcal{A}}
            \Bigl(2\sqrt{N_{t^\pi(B)-1}^\pi(a)}-\frac{(N_{t^\pi(B)-1}^\pi(a))^{3/2}}{2T_{\text{max}}}\Bigl)\Bigl)\\
            &\overset{\underset{(b)}{}}{\leq} \sqrt{L}\Bigl(\frac{K}{\sqrt{\nu}}+ 2\sqrt{KT_{\text{max}}}-\frac{1}{3}\sqrt{\frac{T_{\text{max}}}{K}}\Bigl),
\end{split}
\end{equation*}
where we have utilized that (a) the concaivity of $\sqrt{\cdot}$, and (b) $\min\{\sum_{a=1}^K n_a^{3/2};\sum_{a=1}^K c_a n_a \leq B\} \leq \sum_{a=1}^K (T_{\text{max}}/K)^{3/2} = \sqrt{T_{\text{max}}^3/K}$.
    
Therefore, combining all above result, we obtained the desired result:
\begin{equation*}
\begin{split}
             W^\text{IRS.FH}(B,y) - V(\pi,B,y) \leq&  2\sqrt{L/\nu}  + 2\sqrt{2\log T_{\text{max}}} \times \sum_{b=1}^{B} \frac{\triangle_{t^\pi(b)}^\pi(A_{t^\pi(b)}^\pi,\left\lfloor\frac{B-b_{t^\pi(b)}(A_{1:t^\pi(b)}^\pi)}{c_a}\right\rfloor)}{c_{A_{t^\pi(b)}^\pi}}\\
            \leq&  2\sqrt{L/\nu}  +  2\sqrt{2\log T_{\text{max}}} \times \sqrt{L}\Bigl(\frac{K}{\sqrt{\nu}}+ 2\sqrt{KT_\text{max}}-\frac{1}{3}\sqrt{\frac{T_\text{max}}{K}}\Bigl)\\
            \leq&  2\sqrt{L} \Bigl[\frac{1}{\sqrt{\nu}}+  \sqrt{2\log T_{\text{max}}} \times \Bigl(\frac{K}{\sqrt{\nu}}+ 2\sqrt{K T_\text{max}}-\frac{1}{3}\sqrt{\frac{T_\text{max}}{K}}\Bigl)\Bigl].
\end{split}
\end{equation*}

\subsection{IRS.V-Zero}
    
Let redefine $Q_t^{in}( a_{1:t-1}, a, \omega; B, y )$ as the Q-value associated with the inner problem being solved by IRS.V-Zero, and let 
$$A_t^\star \triangleq \begin{cases}
argmax_{a \in \mathcal{A}} Q_t^{in}( A_{1:t-1}^\pi, a, \omega; B, y ) \quad &\text{if } t <\tau(A_{1:T_{max})} \\
A_{\text{null}} &\text{otherwise}
\end{cases}.$$
Then, the suboptimality gap can be decomposed
\begin{equation*}
\begin{split}
        W^\text{IRS.V-Zero}(B,y) - V(\pi,B,y) 
        	&= \mathbb{E}_y\Bigl[\sum_{t=1}^{T_\text{max}} \max_{a}\{Q_{t}^{in}(A_{1:t-1}^{\pi},a,\omega;B,y)\} - Q_{t}^{in}(A_{1:t-1}^{\pi},A_t^{\pi},\omega;B,y) \Bigl]
        \\&\leq \mathbb{E}_y\Bigl[\sum_{t=1}^{T_\text{max}}
        		\max_{0 \leq n \leq \left\lfloor (B-b_{t-1}(A_{1:t-1}^\pi))/c_{A_t^\star} \right\rfloor }\hat{\mu}_t^\pi(A_t^{\star},n) -\hat{\mu}_t^\pi(A_t^\pi,0)\Bigl]
        \\&\leq \mathbb{E}_y\Bigl[\sum_{b=1}^{B} 
        \max_{0 \leq n \leq \left\lfloor (B-b)/c_{A_t^\star}\right\rfloor } \frac{\hat{\mu}_{t^\pi(b)}^\pi(A_{t^\pi(b)}^\star,n)}{c_{A_{t^\pi(b)}^\star}} - \frac{\hat{\mu}_{t^\pi(b)}^\pi(A_{t^\pi(b)}^\pi,0)}{c_{A_{t^\pi(b)}^\pi}}\Bigl].
\end{split}
\end{equation*}
Define the sequence of confidence intervals:
\[L_{t}(a) \triangleq  \hat{\mu}_{t}^\pi(a,0) , \qquad  U_{t}(a) \triangleq  \hat{\mu}_{t}^\pi(a,0) + \sqrt{2\log (\frac{B}{c_a})} \times \triangle_{t}^\pi(a,\left\lfloor\frac{B-b_{t-1}(A_{1:t-1}^\pi)}{c_a}\right\rfloor)\]
By Lemma 6 in \cite{min2019thompson}, we have
\begin{equation*}
\begin{split}
        \mathbb{E}\Bigl[(\max_{0 \leq n \leq \left\lfloor (B-b_{t-1}(A_{1:t-1}^\pi))/c_{A_t^\star} \right\rfloor }\hat{\mu}_t^\pi(a,n)- U_t(a))^+ | \mathcal{F}_{t-1} \Bigl] \leq& \frac{\triangle_{t}^\pi(a,\left\lfloor\frac{B-b_{t-1}(A_{1:t-1}^\pi)}{c_a}\right\rfloor)}{\sqrt{2\log (\frac{B}{c_a})}} e^{-\frac{2\log (\frac{B}{c_a})}{2}}
        \\ \leq&\frac{\triangle_{t}^\pi(a,\infty)}{\sqrt{2\log (\frac{B}{c_a})}} e^{-\frac{2\log (\frac{B}{c_a})}{2}} \leq \frac{\sqrt{L/\nu}}{B/c_{a}}.
\end{split}     \end{equation*}
where $\mathbb{E}\Bigl[\hat{\mu}_t^\pi(a,0) - L_t(a) |\mathcal{F}_{t-1} \Bigl] = 0$.
The he rest of the proof is almost similar to \ref{sec:subopt of FH}.
We can get an upper bound on the suboptimality gap of IRS.V-Zero as
\begin{equation*}
\begin{split}
\begin{split}
W^\text{IRS.V-Zero}(B,y) - V(\pi,B,y)  \leq& \sqrt{L/\nu}  +  \sqrt{2\log T_{\text{max}}}\times \mathbb{E}_y \Bigl[ \sum_{b=1}^{B}\frac{\triangle_{t^\pi(b)}^\pi(A_{t^\pi(b)}^\pi,\left\lfloor\frac{B-b_{t^\pi(b)}(A_{1:t^\pi(b)}^\pi)}{c_a}\right\rfloor)}{c_{A_{t^\pi(b)}^\pi}}\Bigl]\\
    	   \leq& \sqrt{L/\nu}  + \sqrt{2\log T_{\text{max}}} \times \sqrt{L}\Bigl(\frac{K}{\sqrt{\nu}}+ 2\sqrt{KT_{\text{max}}}-\frac{1}{3}\sqrt{\frac{T_{\text{max}}}{K}}\Bigl)\\
 	   =& \sqrt{L} \Bigl[\frac{1}{\sqrt{\nu}}+ \sqrt{2\log T_{\text{max}}}\times \Bigl(\frac{K}{\sqrt{\nu}}+ 2\sqrt{KT_{\text{max}}}-\frac{1}{3}\sqrt{\frac{T_{\text{max}}}{K}}\Bigl)\Bigl].
\end{split}
\end{split}
\end{equation*}
\section{Random cost setting}\label{sec:random numerical}

\begin{figure*}[htp!]
	\centering
	\includegraphics[width=0.65\textwidth]{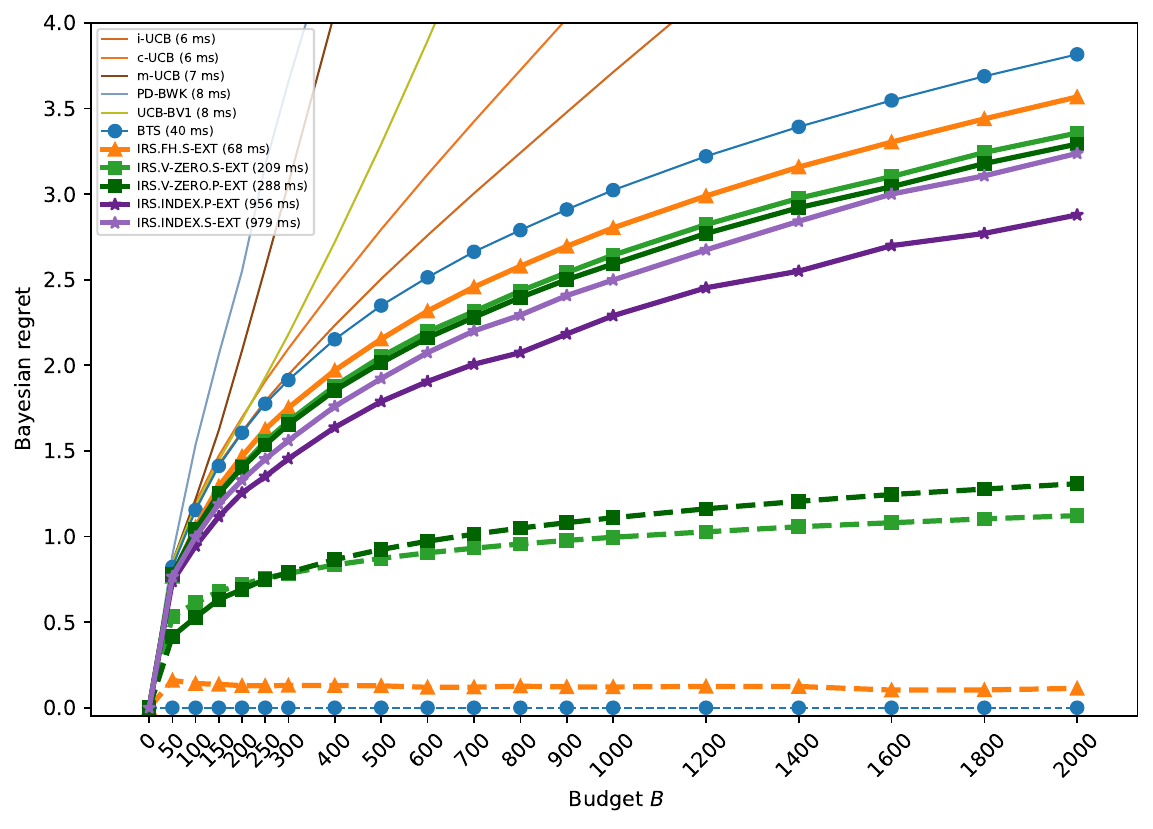}
	\caption{Simulation result in the Beta-Bernoulli MAB with two arms incurring random costs.}
   	\label{fig:simulation_random}
\end{figure*}

We show that IRS algorithms are sufficiently scalable for random cost setting through numerical simulation.
We compare Algorithm \ref{alg:s-ext-irs.fh}, \ref{alg:s-ext-irs.vzero}, \ref{alg:p-ext-irs.vzero}, two extensions of IRS.INDEX policy (Algorithm \ref{alg:irs.index.p-ext} in Appendix \ref{subsec:p-ext}), and all benchmarks discussed in \S 6 except KUBE.
We examine a Beta-Bernoulli MAB instance with $K=2$, and $\alpha_a^{r} = \beta_a^{r}=\alpha_a^{c} = \beta_a^{c} = 1, \forall a \in \mathcal{A}$, where $\mathbb{P}[ C_{a,n} = 20 ] = \theta_a^c = 1 - \mathbb{P}[ C_{a,n} = 10 ]$ with $\theta_a^c \sim \text{Beta}( \alpha_a^{c}, \beta_a^{c} )$.
Figure \ref{fig:simulation_random} reports the result of 50,000 runs of simulation.

Similar to the deterministic cost setting, BTS still outperforms the other benchmarks: when $B=2000$, it shows a performance improvement at least by $35\%$ (regret of  i-UCB compared to regret of BTS).
Our proposed policies even further improve BTS: IRS.FH.S-EXT, IRS.V-Zero.S-EXT, IRS.V-Zero.P-EXT, IRS.INDEX.S-EXT, and IRS.INDEX.P-EXT, respectively, achieve 7\%, 12\%, 14\%, 15\% and 25\% improvement over BTS in terms of reduction in regret.
From the regret lower bound $W^\text{BTS} - W^\text{IRS.V-ZERO.P-EXT}$, we can deduce that no policy can achieve an improvement more than 66\%.

This result shows better performance and tighter bounds in the order of BTS, IRS.FH, IRS.V-Zero, and IRS.INDEX similar to the deterministic cost setting. Also, comparing simple extensions and extensions with additional penalties, if more penalty is imposed instead of sampling more information, it has better performance and tighter bounds. This is a characteristic of IRS frameworks and means that IRS frameworks are sufficiently scalable with random cost setting.

\end{document}